\documentclass[]{fairmeta}

\usepackage[T1]{fontenc}
\usepackage[utf8]{inputenc}
\usepackage{hyperref}
\usepackage{url}
\usepackage{booktabs}
\usepackage{multirow}
\usepackage{makecell}
\usepackage{amsfonts}
\usepackage{amsmath}
\usepackage{amssymb}
\DeclareMathOperator*{\argmin}{arg\,min}
\usepackage{nicefrac}
\usepackage{microtype}
\usepackage{xcolor}
\definecolor{navy}{RGB}{0,0,128}
\usepackage{graphicx}
\usepackage{caption}
\makeatletter
\long\def\@makefntext#1{\noindent\hb@xt@1.8em{\hss\@makefnmark}#1}
\makeatother
\usepackage{enumitem}
\usepackage{chngcntr}
\usepackage{tikz}
\usetikzlibrary{positioning, fit, arrows.meta, decorations.pathreplacing, calc}
\usepackage{tcolorbox}
\usepackage{twemojis}
\newtcolorbox{findingbox}[1]{
  colback=gray!10,
  colframe=gray!10,
  boxrule=0pt,
  arc=3pt,
  left=8pt, right=8pt, top=3pt, bottom=3pt,
  before skip=6pt,
  after skip=6pt,
  fonttitle=\bfseries,
  title={#1},
  colbacktitle=gray!10,
  coltitle=black,
  detach title,
  before upper={\tcbtitle}
}

\title{MobileMoE: Scaling On-Device Mixture of Experts}

\author{Yanbei Chen}
\author{Hanxian Huang}
\author{Ernie Chang}
\author{Jacob Szwejbka}
\author{Digant Desai}
\author{Zechun Liu}
\author{\\Vikas Chandra}
\author{Raghuraman Krishnamoorthi}

\affiliation{Meta AI}

\abstract{Mixture-of-Experts (MoE) has become the de facto architecture for hundred-billion-parameter language models, yet its advantages at sub-billion scales for on-device deployment remain largely unexplored. To close this gap, we present \textbf{MobileMoE}, a family of on-device MoE language models with sub-billion active parameters (0.3-0.9B active and 1.3-5.3B total) that establish a new Pareto frontier for on-device LLMs. We first formulate an on-device MoE scaling law that jointly optimizes MoE architecture under mobile memory and compute constraints, identifying an on-device sweet spot -- moderate sparsity with fine-grained and shared experts -- that is simultaneously memory and compute-optimal. Building on the derived architectures, we train MobileMoE with a four-stage recipe covering pre-training, mid-training, instruction fine-tuning, and quantization-aware training, all on open-source datasets. Across 14 benchmarks, MobileMoE matches or exceeds leading on-device dense LLMs with 2-4$\times$ fewer inference FLOPs, and matches or surpasses the state-of-the-art MoE OLMoE-1B-7B with up to 60\% fewer parameters. To bridge the last mile to mobile deployment, we provide the first efficient MoE inference on commodity smartphones with comprehensive on-device profiling. At comparable INT4 weight memory, MobileMoE-S delivers $1.8$-$3.8\times$ faster prefill and $2.2$-$3.4\times$ faster decode than the dense baseline MobileLLM-Pro.}

\date{May 27, 2026 (initial release)}
\metadata[Venue]{NeurIPS 2026 (December 2026)}
\metadata[Released model collections]{\twemoji{hugs}~\href{https://huggingface.co/collections/facebook/mobilemoe}{huggingface.co/collections/facebook/mobilemoe}}
\correspondence{Yanbei Chen at \email{yanbeichen@meta.com}\\Detailed author contributions can be found in the \hyperref[sec:contributions]{Author Contributions} section.}

\begin{document}

\maketitle

\section{Introduction}
\label{section:intro}

Mixture-of-Experts (MoE) architectures increasingly dominate state-of-the-art Large Language Models (LLMs), as represented by both open-source models (e.g., DeepSeek V3~\cite{deepseek}, Qwen3 MoE~\cite{qwen3}) and proprietary frontier models (e.g., Gemini~\cite{gemini}, Grok~\cite{grok}). However, on-device LLMs remain overwhelmingly dense (e.g., MobileLLM~\cite{mbllm}, MobileLLM Pro~\cite{mbllmpro}), and scaling MoE in the sub-billion active-parameter regime, where on-device LLMs typically operate, remains largely unexplored. Addressing this gap is increasingly crucial for next-generation edge AI: efficient on-device LLMs reduce reliance on cloud compute and enable low-latency, cost-effective, privacy-preserving applications on smartphones, wearables, and embodied agents.

Unlocking the potential of LLMs on edge devices requires overcoming severe compute and memory constraints. MoE architectures address these constraints through three complementary efficiencies. First, \emph{parameter efficiency}: an MoE model expands total capacity through many expert networks while activating only a sparse fraction per token, matching the performance of a much larger dense counterpart at significantly less inference compute~\cite{qwen3}. Second, \emph{runtime efficiency}: sparse activation reduces inference FLOPs, lowering runtime latency and conserving mobile battery life. Third, \emph{learning efficiency}: expert networks specialize across distinct domains (e.g., knowledge, code, math)~\cite{shazeer2017outrageously}, packing broad multi-task capability into one unified model. Crucially, the recent growth of smartphone DRAM in the past few years (e.g., from 4\,GB on iPhone~13 to 12\,GB on iPhone~17, from 8\,GB on Samsung Galaxy S21 to 12\,GB, 16\,GB on S25 and S25 Ultra) provides the memory headroom to host these efficient and capable sparse LLMs directly on mobile devices.

Yet the scaling methodology of on-device MoE, from architectures to training recipes and practical on-device deployment, has yet to be established. While scaling laws have long served as the north star guiding the development of dense LLMs~\cite{kaplan2020scaling,hoffmann2022chinchilla} and MoEs~\cite{clark2022unified,krajewski2024scaling}, existing frameworks overwhelmingly focus on scaling models up to tens or hundreds of billions of parameters for deployment on cloud servers. To address practical edge constraints, we formulate a novel MoE scaling law tailored to the sub-billion active-parameter regime, providing a principled foundation to guide architectural design under joint memory and compute constraints. Building upon this scaling law, we derive \textbf{MobileMoE}, the first sub-billion-active MoE family optimized for the edge across three scales (S/M/L): 0.3B/0.5B/0.9B active parameters (1.3B/2.8B/5.3B total) with $<$3\,GB INT4 weight footprints to fit in mobile DRAM.

\begin{figure}[!t]
\centering
\begin{minipage}[t]{0.47\textwidth}
\centering
\includegraphics[width=\linewidth]{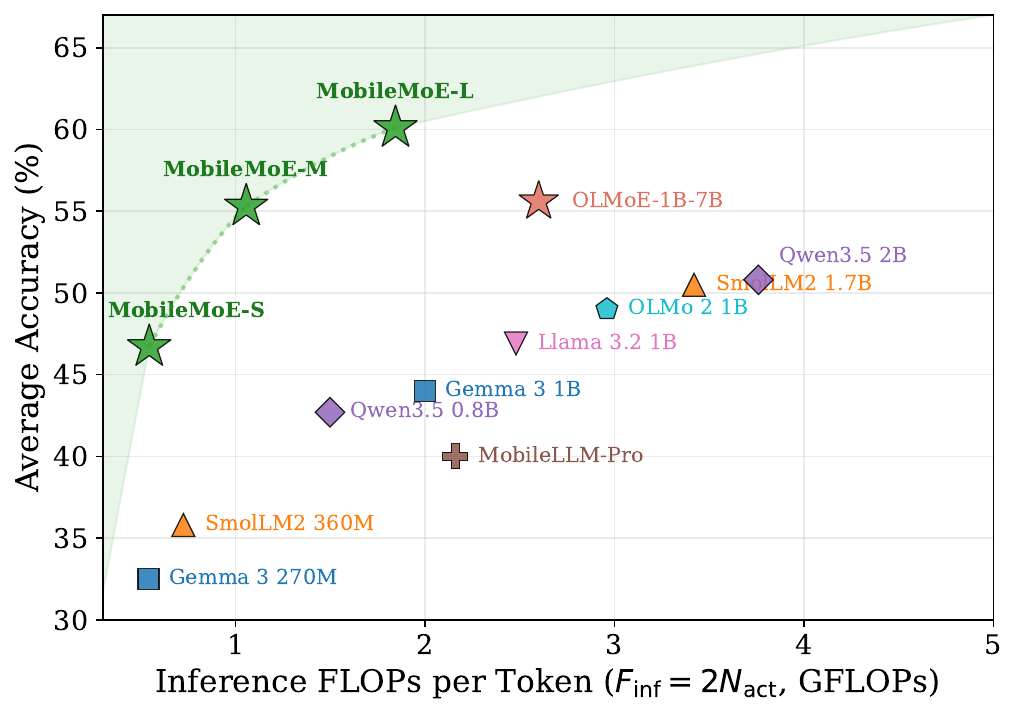}\\
{\small (a)}
\end{minipage}%
\hspace{0.05\textwidth}%
\begin{minipage}[t]{0.47\textwidth}
\centering
\includegraphics[width=\linewidth]{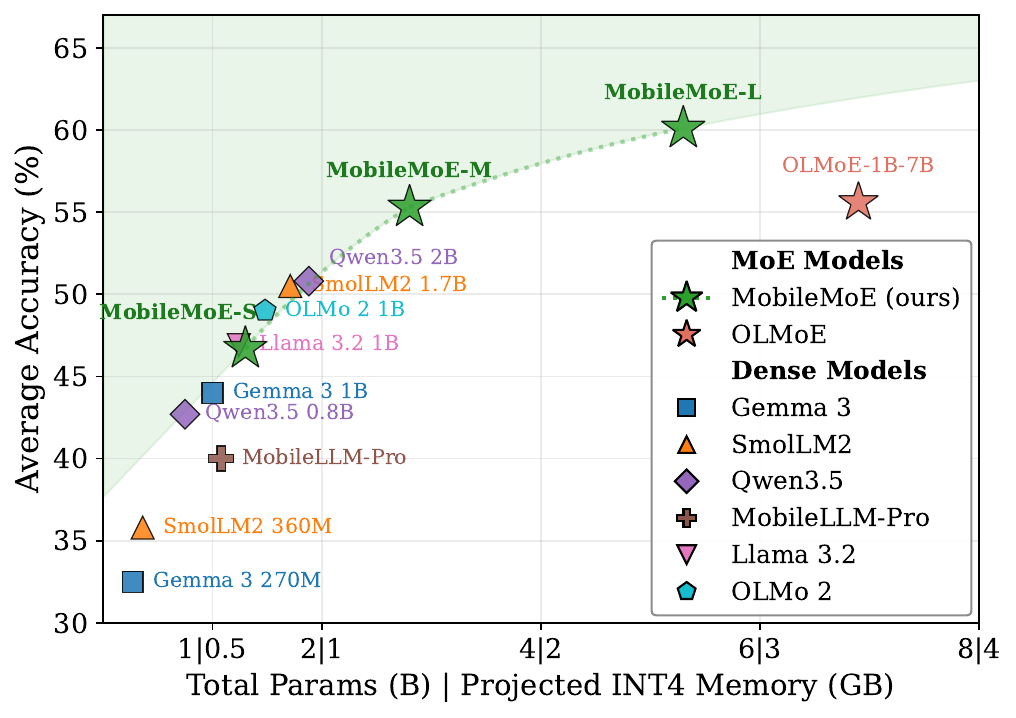}\\
{\small (b)}
\end{minipage}
\caption{\textbf{MobileMoE establishes a new Pareto frontier for on-device LLMs.} Average benchmark accuracy, computed over 14 benchmarks spanning commonsense, knowledge, science, comprehension, and reasoning, is plotted against (a) per-token inference compute $F_{\text{inf}} = 2N_{\text{act}}$ (GFLOPs) and (b) total parameters $N_{\text{total}}$ (B); in (b), x-axis tick labels show total params (B) $|$ projected INT4 memory (GB)\protect\footnotemark.}
\label{fig:pareto}
\end{figure}

To realize the MoE architectural advantages at scale, we design a comprehensive four-stage training recipe: pre-training, mid-training, instruction fine-tuning, and 4-bit quantization-aware training. Our pipeline explicitly addresses MoE-specific training stability and efficiency, and scales up MobileMoE training with exceptional token efficiency. With only $\sim$6T pre-training tokens, MobileMoE matches or surpasses dense baselines trained on 1.5-2$\times$ more tokens (e.g., 9T for Llama~3.2 1B~\cite{llama3}, 11T for SmolLM2 1.7B~\cite{smollm2}), validating the learning efficiency of MoE at the sub-billion active scale. Notably, our scaling-law-derived architecture and training recipe enable MobileMoE to establish a new Pareto frontier for on-device LLMs across 14 foundational benchmarks spanning commonsense, knowledge, science, comprehension, and reasoning (Figure~\ref{fig:pareto}). Smaller MobileMoE-S/M match or exceed dense baselines using 2-4$\times$ fewer inference FLOPs at comparable memory, while MobileMoE-L pushes the frontier further to state-of-the-art accuracy at sub-billion active scale. Furthermore, compared to the state-of-the-art MoE OLMoE-1B-7B~\cite{olmoe}, MobileMoE-M matches its accuracy with $\sim$60\% fewer active and total parameters, while MobileMoE-L achieves much higher accuracy with 30\% fewer active parameters and 23\% smaller model memory footprint.

\footnotetext{Accuracy is reported from public model releases at full precision (BF16). In subplot (b), the x-axis shows total parameters $N_{\text{total}}$ in billions alongside the corresponding projected INT4 weight memory, computed as $\mathcal{M}_{\text{weight}} = \tfrac{1}{2}N_{\text{total}}$\,GB under 4-bit weight quantization that is commonly used for on-device LLM deployment (as detailed in Section~\ref{sec:training_recipes}, Section~\ref{sec:ondevice}).}

Beyond benchmark performance, we demonstrate the practical on-device runtime benefits by deploying MobileMoE on flagship smartphones: Samsung Galaxy~S25 and iPhone~16~Pro. Since most existing mobile inference stacks lack native MoE support, we develop a custom fused MoE kernel to enable efficient MoE inference, providing the first MoE runtime support on commodity smartphone CPUs, with comprehensive runtime profiling across CPU and GPU backends. Powered by this kernel, at comparable INT4 weight memory, MobileMoE-S achieves $1.8$--$3.8\times$ faster prefill and $2.2$--$3.4\times$ faster decode than the dense baseline MobileLLM-Pro~\cite{mbllmpro} while consuming up to 22\% less peak RSS at 8k context. Concurrently, MobileMoE-M matches or outperforms MobileLLM-Pro on runtime with higher accuracy, while MobileMoE-L delivers substantially higher accuracy with moderate runtime cost. The consistency of this Pareto pattern on mobile devices confirms the compute and memory efficiency of MobileMoE holds on real hardware.

Our contributions are three-fold:
\begin{enumerate}
    \item We introduce MobileMoE-S/M/L, the first sub-billion-active MoE family for on-device deployment, derived based on a generalized on-device MoE scaling law under joint memory and compute constraints. Guided by this scaling law, we identify the sweet-spot MoE design choices for on-device use cases (moderate sparsity, fine-grained granularity, and a shared expert) that together define MobileMoE.
    \item We propose a four-stage training recipe (pre-training $\rightarrow$ mid-training $\rightarrow$ SFT $\rightarrow$ INT4 QAT) with MoE-specific stability and efficiency techniques. The recipe scales MobileMoE to Pareto-leading accuracy at only $\sim$6T pre-training tokens -- substantially fewer than dense baselines (9T for Llama~3.2 1B, 11T for SmolLM2), while surpassing the state-of-the-art MoE OLMoE-1B-7B with fewer total parameters.
    \item We deploy MobileMoE on commodity smartphones (Samsung Galaxy~S25, iPhone~16~Pro) via a custom fused MoE kernel in ExecuTorch with systematic runtime profiling. MobileMoE-S achieves $1.8$--$3.8\times$ faster prefill and $2.2$--$3.4\times$ faster decode than the dense MobileLLM-Pro at comparable INT4 weight memory, establishing MoE as a practical path for efficient on-device LLMs.
\end{enumerate}

\section{Related Work}
\label{sec:related}

\textbf{On-Device LLMs} enable fast, privacy-preserving, and cost-effective inference at the edge, but must operate under stringent latency and memory constraints distinct from server-side deployments. A growing body of recent work has introduced dense on-device LLMs at sub-billion to few-billion scales: MobileLLM~\cite{mbllm}, MobileLLM-Pro~\cite{mbllmpro} adopt deep-and-thin architectures to maximize parameter efficiency at sub-billion scales, SmolLM~\cite{smollm2}, Gemma~\cite{team2024gemma,gemma3report2025} provide families of small LLMs with competitive quality, and MobileLLM-Flash~\cite{huang2026mobilellm}, Nemotron-Flash~\cite{fu2025nemotron} use architecture search to optimize on-device latency. Another line of work improves hardware efficiency through NAS methods~\cite{proxylessnas,onceforall,shen2024search,merino,gu2026jet} that optimize architectures for specific deployment targets. These efforts focus exclusively on dense architectures, where scaling model quality inherently demands increasing higher active parameter counts and inference compute. We pursue MoE as a complementary path that expands model capacity at minimal per-token compute for efficient on-device deployment.

\textbf{Mixture of Experts (MoE)} offers a parameter-efficient paradigm by routing tokens to a sparse subset of specialized expert networks~\cite{jacobs1991adaptive,shazeer2017outrageously,zoph2022st,fedus2022switch,zhou2022mixture,aria2024}. Concretely, MoE expands the learning capacity of modern transformers by replacing the dense feed-forward block in each layer with a set of expert subnetworks, increasing total parameters while keeping active parameters compact through sparse routing. Beyond parameter scaling, MoE also enables expert specialization: the routing mechanism learns to assign different token types to dedicated experts, allowing subnetworks to specialize in distinct linguistic tasks~\cite{shazeer2017outrageously} and broader multimodal domains~\cite{aria2024}. By decoupling total parameters from active inference compute, MoE has driven the scaling of state-of-the-art LLMs, e.g., Mixtral~\cite{mixtral}, DeepSeek-MoE~\cite{dai2024deepseekmoe,deepseek}, and Qwen-MoE~\cite{qwen2,qwen3}. While efforts such as OLMoE~\cite{olmoe} have explored smaller scales, the sub-billion active-parameter regime -- where on-device LLMs operate efficiently~\cite{mbllm,mbllmpro} -- remains unexplored under practical edge constraints. Our work specifically studies MoE at this scale, with systematic analyses of architectural choices under on-device constraints.

\textbf{Scaling Laws} characterize power-law relationships between compute, data, and parameters~\cite{kaplan2020scaling,hoffmann2022chinchilla}, providing a principled foundation for LLM development, covering compute-optimal parameter-data allocation~\cite{hoffmann2022chinchilla,llama3}, training hyperparameters~\cite{bi2024deepseek}, learning rate schedules~\cite{hu2024minicpm}, and data mixtures~\cite{shukor2025scaling}. Scaling laws have also been extended to MoE, studying expert count~\cite{clark2022unified}, expert granularity~\cite{krajewski2024scaling}, and expert allocation under memory constraints~\cite{ludziejewski2025joint}. These existing formulations, however, primarily target server-scale LLMs, where abundant hardware resources make large model memory footprints feasible while inference can be parallelized across server GPUs to improve runtime efficiency. By contrast, on-device deployment requires jointly considering inference cost and memory footprint, which are governed by active and total parameters, respectively. While existing scaling laws target server-scale LLMs, we formulate an on-device MoE scaling law to derive architecture under mobile memory and compute constraints, with an end-to-end training recipe to scale sub-billion-active MoE on devices.

\section{Scaling On-Device MoE}

\subsection{Preliminaries}
\label{sec:preliminaries}

\begin{figure*}[t]
\centering
\begin{minipage}[c]{0.43\textwidth}
\centering
\includegraphics[width=\linewidth]{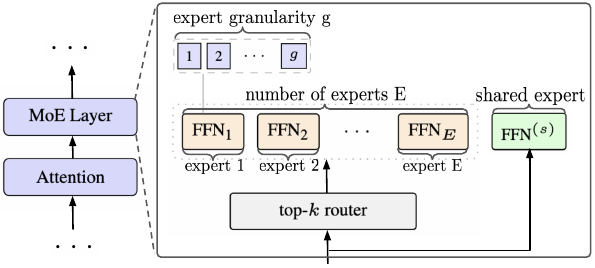}
\end{minipage}%
\hfill
\begin{minipage}[c]{0.55\textwidth}
\centering
\footnotesize
\setlength{\tabcolsep}{2.2pt}
\resizebox{\linewidth}{!}{%
\begin{tabular}{@{}lrrrrrrccc@{}}
\toprule
& \multicolumn{5}{c}{\textbf{Base Arch.}} & \multicolumn{3}{c}{\textbf{MoE Design}} \\
\cmidrule(lr){2-6} \cmidrule(lr){7-9}
\textbf{Model} & $d_{\text{model}}$ & $d_{\text{ff}}$ & $n_h$ & $n_{\text{kv}}$ & $n_l$ & $E$ & $g$ & Shared \\
\midrule
Small (S) & 768 & 3072 & 12 & 4 & 20 & \{1--32\} & \{1--16\} & \{\checkmark, \texttimes\} \\
Medium (M) & 1024 & 4096 & 16 & 4 & 26 & \{1--32\} & \{1--16\} & \{\checkmark, \texttimes\} \\
Large (L) & 1280 & 5120 & 20 & 4 & 32 & \{1--32\} & \{1--16\} & \{\checkmark, \texttimes\} \\
\bottomrule
\end{tabular}
}
\end{minipage}
\caption{\textbf{MobileMoE architectures.} \textit{Left}: MoE model design space with three design factors: model sparsity $(E, k)$, expert granularity $g$, shared experts $s$. \textit{Right}: MobileMoE at three scales: Small (S), Medium (M), Large (L) with 0.3, 0.5, 0.9B active parameters, using a base architecture of expansion ratio $d_{\text{ff}}/d_{\text{model}} = 4$, aspect ratio $d_{\text{model}}/n_l \approx 40$, SwiGLU, GQA with 4 KV heads, along with variants of MoE design choices on number of experts $E \in \{1, 2, 4, 8, 16, 32\}$, fine-grained expert granularity $g \in \{1, 2, 4, 8, 16\}$, and shared expert \{\checkmark, \texttimes\}.}
\label{fig:moe-arch}
\label{tab:mobilemoe-models}
\end{figure*}

\textbf{Mixture-of-Experts (MoE).}
Consider a decoder-only transformer with $n_l$ layers of dimension $d_{\text{model}}$. Each layer consists of grouped-query attention (GQA): $n_h$ query heads, $n_{\text{kv}}$ key-value heads, followed by a feed-forward network (FFN) of hidden dimension $d_{\text{ff}}$. An MoE model replaces the dense FFN with $E$ routed expert FFNs and a top-$k$ router that selects the $k$ highest-scoring experts per token. State-of-the-art MoE models differ widely in architecture choices: DeepSeek-V3~\cite{deepseek} uses 256 fine-grained experts with top-8 routing and a shared expert, Qwen3-MoE~\cite{qwen3} uses 128 experts with top-8 routing but no shared expert, and Mixtral~\cite{mixtral} uses 8 coarse-grained experts with top-2 routing. These differences highlight a lack of consensus at scale on key design choices. Crucially, these choices remain largely under-explored for on-device models, where resource constraints differ fundamentally. We therefore study three factors (Figure~\ref{fig:moe-arch}, left): (i) \textit{model sparsity} $(E, k)$, where routed expert count $E$ and active expert count $k$ control the ratio of active to total parameters; (ii) \textit{expert granularity} $g$, where each routed expert is split into $g$ sub-experts of hidden dimension $d_{\text{ff}}/g$, yielding $gE$ experts with $gk$ activated experts per token; and (iii) \textit{shared expert} $s$, an always-on expert that bypasses routing. Formally, the MoE layer output is $\mathbf{y} = \sum_{i \in \text{Top-}gk} \text{router}_i(\mathbf{x}) \cdot \text{FFN}_i(\mathbf{x}) + \text{FFN}^{(s)}(\mathbf{x})$.

\textbf{On-Device LLMs.} Existing LLMs follow practical rules of thumb in model design, e.g., GPT-3~\cite{gpt3} uses FFN expansion ratio $d_{\text{ff}}/d_{\text{model}} = 4$ and width-depth aspect ratio $d_{\text{model}}/n_l = 128$, while on-device LLMs (e.g., MobileLLM~\cite{mbllm}, MobileLLM Pro~\cite{mbllmpro}) adopt a smaller aspect ratio of approximately 40, favoring deeper architectures in the sub-billion-parameter regime. Building on this principle, we instantiate our on-device MoE models with a base backbone defined by $d_{\text{ff}}/d_{\text{model}} = 4$, $d_{\text{model}}/n_l \approx 40$, 4 key-value heads, and optimize MoE-specific choices (Figure~\ref{tab:mobilemoe-models}, right) using our on-device scaling law.

\subsection{On-Device MoE Scaling Law}
\label{sec:scaling-law}

Unlike server-side deployments with abundant resources, on-device use cases (e.g., smartphones, wearables) face strict hardware constraints, requiring explicit trade-offs among performance, model size, and inference cost. We navigate these trade-offs systematically by jointly optimizing MoE architectures under device memory and compute constraints, over the design space in Figure~\ref{fig:moe-arch}.

\textbf{On-Device MoE Scaling Law.} Formally, we introduce a generalized on-device MoE scaling law:
\begin{equation}
    \mathcal{L}(N_{\text{act}}, D, \hat{E}, x) = A_x \hat{E}^{\delta_x} N_{\text{act}}^{\alpha_x + \gamma_x \ln \hat{E}} + B_x \hat{E}^{\omega_x} D^{\beta_x + \zeta_x \ln \hat{E}} + c_x
    \label{eq:generalized-moe}
\end{equation}
where $\mathcal{L}$ is the model loss, $N_{\text{act}}$ is active parameters, $D$ is training data size, $\hat{E}$ is a monotonic transformation of the number of expert $E$ which decides the total parameters $N_{\text{total}}$ and the model sparsity (i.e., $\text{sparsity}=1-N_{\text{act}}/N_{\text{total}}$), and $x$ refers to architecture choices: expert granularity $g$, shared experts $s$, which does not necessarily change the parameters $N_{\text{act}}, N_{\text{total}}$, and $c_x$ is the irreducible loss.
This formulation admits two {\em reduced forms} that recover established scaling laws as special cases.

\textit{Reduced form I} ($\mathcal{L}|_{x}$). With architecture choice $x$ fixed, Eq.~\eqref{eq:generalized-moe} reduces to \textit{joint MoE scaling law}~\cite{ludziejewski2025joint}:
\begin{equation}
    \mathcal{L}_x(N_{\text{act}}, D, \hat{E}) = A \hat{E}^{\delta} N_{\text{act}}^{\alpha + \gamma \ln \hat{E}} + B \hat{E}^{\omega} D^{\beta + \zeta \ln \hat{E}} + c
    \label{eq:joint-moe}
\end{equation}
which absorbs $x$ as a constant: $A = A_x$, $\alpha = \alpha_x$, $\delta = \delta_x$, $\gamma = \gamma_x$, $B = B_x$, $\beta = \beta_x$, $\omega = \omega_x$, $\zeta = \zeta_x$, $c = c_x$. This reduced form was derived to find \emph{memory-optimal} expert counts in MoE, where $\hat{E}$ is a monotonic transformation of the number of experts $E$ defined as $\frac{1}{\hat{E}} = \frac{1}{E - 1 + \left(\frac{1}{E_{\text{start}}} - \frac{1}{E_{\text{max}}}\right)^{-1}} + \frac{1}{E_{\text{max}}}$~\cite{clark2022unified}.

\textit{Reduced form II} ($\mathcal{L}|_{\hat{E}}$). With expert count $E$ fixed, Eq.~\eqref{eq:generalized-moe} reduces to \textit{Chinchilla scaling law}~\cite{hoffmann2022chinchilla}:
\begin{equation}
    \mathcal{L}_{\hat{E}}(N_{\text{act}}, D, x) = \tilde{A}_x N_{\text{act}}^{\tilde{\alpha}_x} + \tilde{B}_x D^{\tilde{\beta}_x} + c_x
    \label{eq:chinchilla}
\end{equation}
which absorbs $\hat{E}$ as constants: $\tilde{A}_x = A_x \hat{E}^{\delta_x}$, $\tilde{\alpha}_x = \alpha_x + \gamma_x \ln \hat{E}$, $\tilde{B}_x = B_x \hat{E}^{\omega_x}$, $\tilde{\beta}_x = \beta_x + \zeta_x \ln \hat{E}$. This reduced form is equivalent to standard scaling laws for finding \emph{compute-optimal} architecture choices.

\textbf{On-Device Optimization Objective.} For on-device deployment, the optimization of model architecture includes both compute and memory constraints; thus, we minimize Eq.~\eqref{eq:generalized-moe} subject to
\begin{equation}
\begin{aligned}
    \argmin_{N_{\text{act}}, D, E, x} \quad & \mathcal{L}(N_{\text{act}}, D, \hat{E}, x) \\
    \text{s.t.} \quad & \text{compute:} \;\; F_{\text{train}} = 6 N_{\text{act}} D \; \text{(training)};  \;\; F_{\text{inf}} = 2 N_{\text{act}} \; \text{(per-token inference)} \\
    & \text{memory:} \;\; \mathcal{M}(N_{\text{total}}, T) \leq M \; \text{(device memory budget)}
\end{aligned}
\label{eq:opt-objective}
\end{equation}
where $F_{\text{train}}$ is the training compute budget in FLOPs, $F_{\text{inf}}$ is the per-token inference compute in FLOPs (forward pass only), and $M$ is the device DRAM budget in GB, roughly capped at 5\,GB for app usage on current smartphones. The memory function $\mathcal{M}$ accounts for both total parameters $N_{\text{total}}$ and the KV cache at context length $T$. Prior work on on-device LLMs has demonstrated that low-bit quantization (e.g., 4-bit weights, 8-bit KV cache) substantially reduces memory footprint while retaining model quality~\cite{mbllmpro,Fedorovetal2024}. Following this practice, we formulate the memory function as
\begin{equation}
    \mathcal{M}(N_{\text{total}}, T) = \underbrace{\tfrac{b_w}{8} N_{\text{total}}}_{\mathcal{M}_{\text{weight}}} + \underbrace{\tfrac{b_{\text{kv}}}{8} \cdot 2 T n_l n_{\text{kv}} d_h}_{\mathcal{M}_{\text{KV cache}}}, \text{with} \; \mathcal{M}(N_{\text{total}}, T) \leq M
    \label{eq:memory}
\end{equation}
where ${\mathcal{M}_{\text{weight}}}$ is the quantized model weight memory with $b_w$-bit precision (e.g., 4 for INT4), and ${\mathcal{M}_{\text{KV cache}}}$ is the KV cache memory with $b_{\text{kv}}$-bit precision (e.g., 8 for INT8), $T$ is the context length, $d_h$ is the head dimension, $\mathcal{M}(N_{\text{total}}, T)$ is a tractable {\em proxy} for the on-device memory required to host the model (static model weights and KV cache, persistent throughout inference), excluding transient activation buffers and runtime overhead which can be optimized via runtime techniques. $M$ is the on-device memory budget.

\subsection{Finding the Optimal On-Device MoE}
\label{sec:finding_optimal}

The on-device MoE scaling law (Eq.~\eqref{eq:generalized-moe}) and optimization objective (Eq.~\eqref{eq:opt-objective}) serve as the principled foundation to govern our MobileMoE architecture design under compute ($F_{\text{train}}$, $F_{\text{inf}}$) and memory ($\mathcal{M}$) constraints. A na\"ive joint sweep over the three design axes in Figure~\ref{fig:moe-arch} (number of experts $E$, expert granularity $g$, and shared expert $s$) would incur a combinatorial number of ablation runs, but these axes are \emph{structurally decoupled} at fixed active parameters $N_{\text{act}}$: $E$ alone changes $N_{\text{total}}$ (and thus memory), $g$ changes the expert networks yet preserves both $N_{\text{act}}$ and $N_{\text{total}}$, and $s$ adds a shared dense pathway, where the shared expert can be sized to retain both $N_{\text{act}}$ and $N_{\text{total}}$ -- so the memory and compute-optimal $E$ is preserved under any subsequent choice of $g$ or $s$. We therefore adopt a divide-and-conquer approach, decomposing the architecture optimization into three controlled ablation studies, each isolating one factor while holding the others fixed, to progressively determine the optimal on-device MoE architecture grounded in the on-device MoE scaling law (Section~\ref{sec:scaling-law}).

\begin{figure}[!t]
\centering
\includegraphics[width=\linewidth]{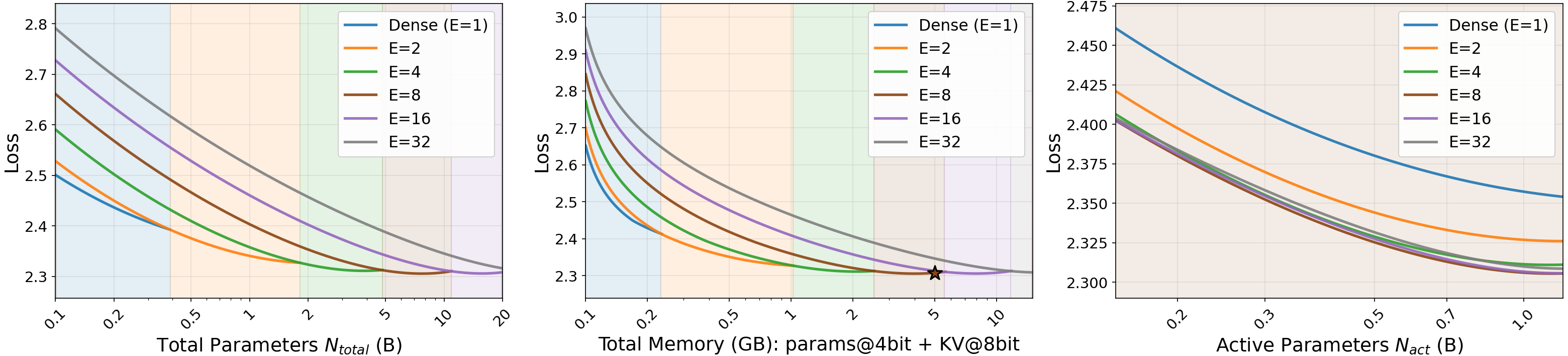}\\
\makebox[0.333\linewidth]{\small (a)}%
\makebox[0.333\linewidth]{\small (b)}%
\makebox[0.333\linewidth]{\small (c)}\\
\caption{\textbf{Scaling the number of experts $E$.} Model loss predicted by the fitted on-device scaling law (Eq.~\eqref{eq:generalized-moe}) plotted against (a) total parameters $N_{\text{total}}$, (b) on-device memory $\mathcal{M}$ (Eq.~\eqref{eq:memory}), and (c) active parameters $N_{\text{act}}$, under a training budget of $5 \times 10^{20}$ FLOPs (our experimental regime). Each curve corresponds to one expert count $E$, truncated at the crossover with the next-larger $E$ curve. Shaded regions indicate the optimal $E$ along each axis; e.g., at $\mathcal{M}=5$\,GB, the optimal $E$ is 8, marked by the star in (b).}
\label{fig:scaling-experts}
\end{figure}

\begin{figure}[!t]
\centering
\includegraphics[width=\linewidth]{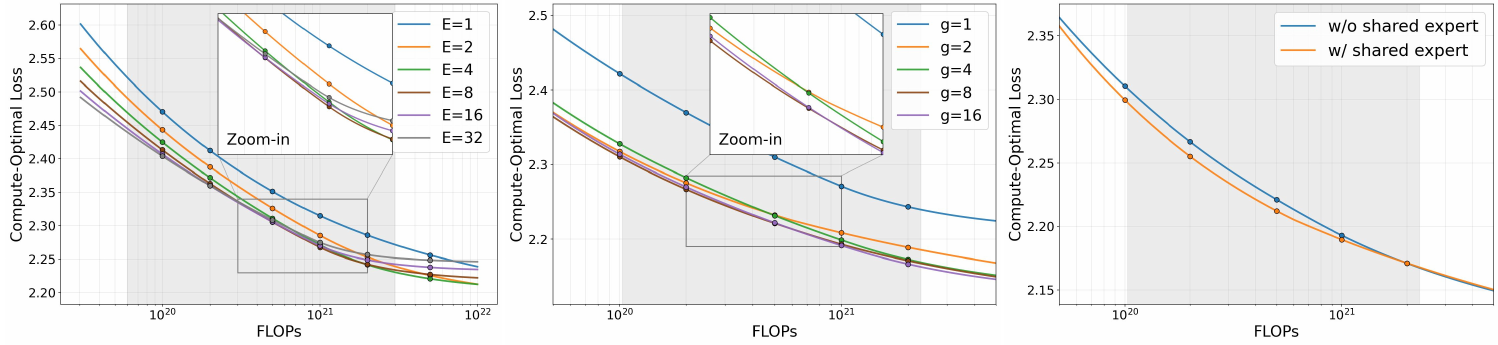}\\
\makebox[0.333\linewidth]{\small (a)}%
\makebox[0.333\linewidth]{\small (b)}%
\makebox[0.333\linewidth]{\small (c)}\\
\caption{\textbf{Scaling MoE models with compute optimality.} The compute-optimal loss vs.\ training compute FLOPs for each MoE design factor predicted by the on-device scaling law (Eq.~\eqref{eq:generalized-moe}): when varying (a) number of experts $E$, (b) expert granularity $g$, (c) with vs without shared expert, the curves with lower model loss indicate the more compute-efficient design choices given a fixed compute budget. Grey area is the experimental regime.}
\label{fig:moe-optimal-flops}
\end{figure}

\textbf{Scaling the number of experts $E$.} Given fixed active parameters $N_{\text{act}}$, the number of experts $E$ determines the total parameters $N_{\text{total}}$, which changes the model sparsity ($1-N_{\text{act}}/N_{\text{total}}$) and model memory (Eq.~\eqref{eq:memory}). To explore the optimal $E$ given fixed device memory constraint, our scaling study sweeps over $E \in \{1, 2, 4, 8, 16, 32\}$ across the three base architectures in Figure~\ref{fig:moe-arch}, spanning the sub-billion active parameter regime $N_{\text{act}} \in \{0.3\text{B}, 0.5\text{B}, 0.9\text{B}\}$ (with the largest $(E, N_{\text{act}})$ combinations exceeding the 5\,GB budget); each model is trained across data budgets $D \in \{100, 200, ..., 500\}$ billion tokens. We fit the on-device scaling law (Eq.~\eqref{eq:generalized-moe}) on these sweep runs with architecture choice $x$ held fixed, and solve the optimization objective (Eq.~\eqref{eq:opt-objective}). Similar to~\cite{hoffmann2022chinchilla}, the scaling coefficients are fitted using LBFGS optimization (detailed in Appendix~\ref{app:fitting}), which provides the scaling curves in Figure~\ref{fig:scaling-experts}, Figure~\ref{fig:moe-optimal-flops}(a), and the following finding.
\begin{findingbox}{\textcolor{navy}{Finding 1:} \mdseries\color{black} MoE models can be both memory-optimal and compute-optimal over dense models. With fixed memory ($M>0.25$\,GB), MoE models ($E>1$) achieve lower loss than dense models (Figure~\ref{fig:scaling-experts}(b)). With fixed compute $F_{\text{inf}}$ (fixed $N_{\text{act}}$) and $F_{\text{train}}$, increasing $E$ reduces loss with diminishing returns beyond $E=8$ (Figure~\ref{fig:scaling-experts}(c), Figure~\ref{fig:moe-optimal-flops}(a)). While the optimal $E$ grows with more memory, moderate sparsity ($E \in \{4, 8\}$) is the practical sweet spot in the on-device memory regime.}
\end{findingbox}

Based on Finding 1, we construct the on-device MoE with $E=8$, denoted as $\mathrm{MoE}(E=8)$, which achieves near-optimal performance at fixed inference compute with sub-billion active parameters (Figure~\ref{fig:scaling-experts}(c)), while remaining within the practical sweet spot under on-device memory constraints (e.g., $\leq$5\,GB in Figure~\ref{fig:scaling-experts}(b)).

\textbf{Scaling the expert granularity $g$.} Varying expert granularity divides each expert into $g$ fine-grained sub-experts while keeping both total and active parameters intact; thus, on-device memory and inference compute remain constant when scaling $g$. Intuitively, finer granularity enables more flexible expert combinations during routing -- with $g \cdot E$ fine-grained experts and top-$k \cdot g$ routing, the router can compose more diverse expert combinations, leading to more specialized routing paths~\cite{dai2024deepseekmoe}. To find the compute-optimal $g$ under the on-device scaling law (Eq.~\eqref{eq:generalized-moe}), our scaling study sweeps over $g \in \{1, 2, 4, 8, 16\}$ upon the model $\mathrm{MoE}(E=8)$ (derived in Finding 1) under the same experimental regime of $N_{\text{act}}$ and $D$, and solves for the compute-optimal $g$ (Figure~\ref{fig:moe-optimal-flops}(b)).
\begin{findingbox}{\textcolor{navy}{Finding 2:} \mdseries\color{black} Fine-grained experts ($g>1$) achieve substantially lower loss at fixed compute (Figure~\ref{fig:moe-optimal-flops}(b)) due to more diverse top-$k$ routing, but exhibit diminishing returns beyond $g=8$.}
\end{findingbox}

Following Finding 2, we adopt a compute-optimal granularity of $g=8$ for $\mathrm{MoE}(E=8)$, which results in $\mathrm{MoE}(E=8, g=8)$, featuring $64$ fine-grained experts and top-$8$ routing. Crucially, this fine-grained expert segmentation maintains the same memory footprint, remaining within on-device limits.

\textbf{Scaling with shared expert $s$.} Whether to incorporate a shared expert -- a dense pathway activated on every token -- remains an open design choice: it is adopted in DeepSeekMoE~\cite{dai2024deepseekmoe} and Qwen2MoE~\cite{qwen2}, yet omitted in OLMoE~\cite{olmoe} and Qwen3MoE~\cite{qwen3}. To isolate the architectural effect of the shared expert, we compare $\mathrm{MoE}(E=8, g=8)$ with and without a shared expert by replacing 4 of the 8 active routed experts with the shared expert ($4\times$ the size of a routed fine-grained expert), yielding 60 routed experts with top-$4$ routing and one shared expert. This specific configuration ensures the routed expert count remains divisible by the expert-parallel size ($\mathrm{EP}=4$), thereby preserving training efficiency. Notably, it also preserves active and total parameters (and thus memory), enabling a fair ablation on the architectural impact of the shared expert. We fit the on-device scaling law (Eq.~\eqref{eq:generalized-moe}) on sweep runs with $E$ and $g$ fixed, under the same experimental regime of $N_{\text{act}}$ and $D$ to identify the optimal setting of $s$ (Figure~\ref{fig:moe-optimal-flops}(c)).
\begin{findingbox}{\textcolor{navy}{Finding 3:} \mdseries\color{black} With a shared expert, the on-device MoE model achieves lower loss than its counterpart without a shared expert given fixed compute FLOPs (Figure~\ref{fig:moe-optimal-flops}(c)).}
\end{findingbox}

Guided by Finding 3, the shared expert (generalist) complements routed experts (specialists). We adopt the shared expert to derive our \textbf{MobileMoE} architecture: $\mathrm{MoE}(E=8, g=8, s=\checkmark)$ -- $60$ fine-grained experts, top-$4$ routing and a shared expert. Applying this to the three base architectures in Figure~\ref{fig:moe-arch}, we obtain \textbf{MobileMoE-S/M/L} with $\{0.3, 0.5, 0.9\}$\,B active parameters and $\{1.26, 2.82, 5.33\}$\,B total parameters, all fitting within a 3-5\,GB on-device memory budget under 4-bit quantization (Eq.~\eqref{eq:memory}).

\begin{figure}[!t]
    \centering
    \includegraphics[width=\linewidth]{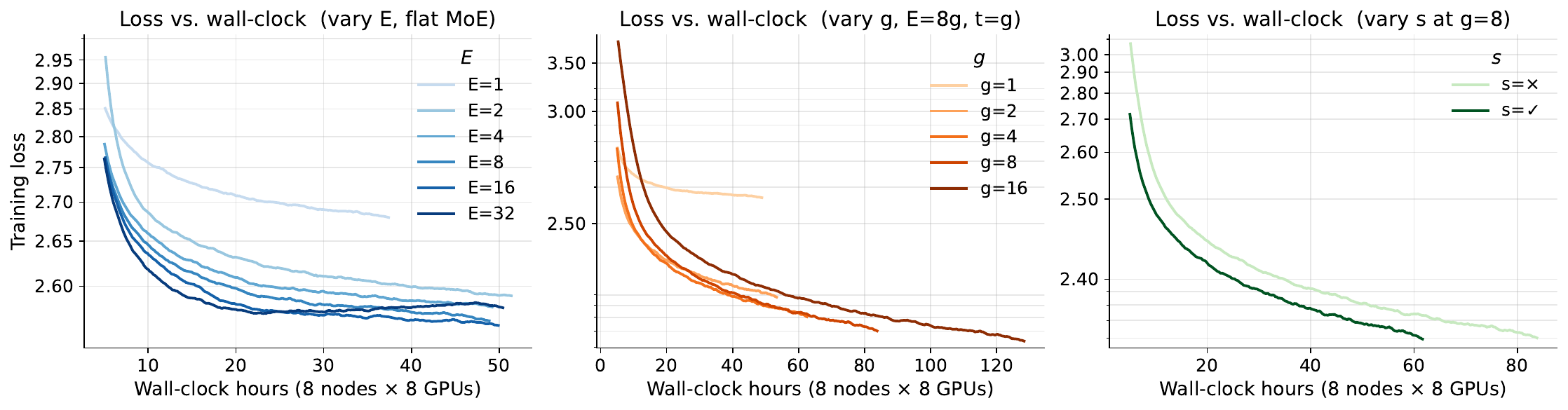}\\
    \makebox[0.333\linewidth]{\small (a)}%
    \makebox[0.333\linewidth]{\small (b)}%
    \makebox[0.333\linewidth]{\small (c)}\\
    \caption{\textbf{Training efficiency of MoE architecture} ($N_{\text{act}}=0.3$B). Training loss vs.\ wall-clock hours when varying (a) number of experts $E \in \{1, 2, 4, 8, 16, 32\}$, (b) expert granularity $g \in \{1, 2, 4, 8, 16\}$ at $E=8$, and (c) with vs.\ without a shared expert $s \in \{\checkmark, \times\}$ at $E=8, g=8$. For fair comparison, each run is trained on the same data for 500B tokens, with identical hardware (8 nodes, 64 NVIDIA H100 96\,GB GPUs) -- see Table~\ref{tab:ablations} for full ablation details.}
    \label{fig:loss-wallclock}
\end{figure}

\textbf{Training efficiency of MoE architecture.} Figure~\ref{fig:loss-wallclock} shows training loss versus wall-clock time for the three design factors, complementing the memory and compute-optimal analysis from a training-efficiency perspective. For model sparsity (Figure~\ref{fig:loss-wallclock}(a)), the dense baseline ($E=1$) trains fastest per step but converges to higher loss, while all MoE configurations ($E\geq 2$) share roughly the same training throughput. The final loss exhibits diminishing returns with $E\geq 8$, where $E=8$ and $E=16$ converge to similar final loss, but $E=32$ shows performance regression despite having more total parameters and memory footprint. Taking into account the higher memory cost at larger $E$ (Eq.~\eqref{eq:memory}), $E=8$ remains a memory- and training-efficient operating point. For expert granularity (Figure~\ref{fig:loss-wallclock}(b)), finer-grained experts ($g \geq 2$) achieve lower loss than $g=1$ (no fine-grained experts) at the same wall-clock budget, while $g=16$ incurs much higher per-step overhead ($\sim$50\% more wall-clock time) than $g=8$ with negligible loss reduction ($<\!0.01$), making $g=8$ a training-efficient sweet spot for expert granularity. For the shared expert (Figure~\ref{fig:loss-wallclock}(c)), adding the shared expert ($s=\checkmark$) further improves training efficiency over routed-only experts ($s=\times$), achieving higher training throughput with lower final loss at the same $N_{\text{act}}$ and $N_{\text{total}}$. Together, these results confirm that the MobileMoE configuration ($E=8, g=8, s=\checkmark$) is also training-efficient on real hardware.

\subsection{Scaling MobileMoE with Full Training Recipe}
\label{sec:training_recipes}

\begin{figure}[!t]
\centering
\includegraphics[width=0.85\linewidth]{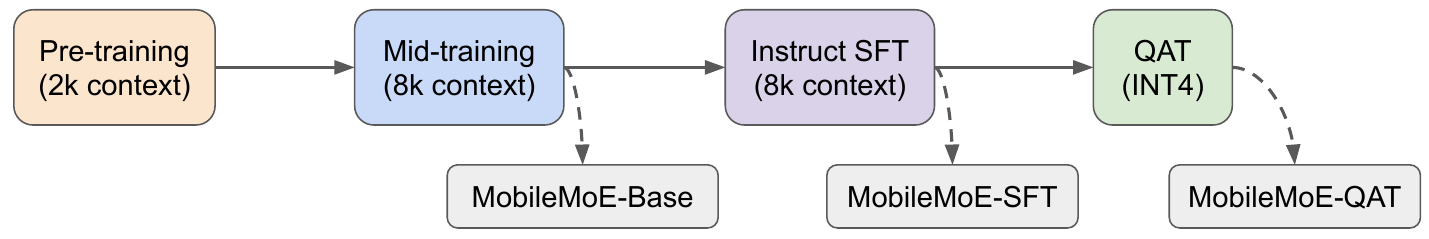}
\caption{\textbf{MobileMoE four-stage training recipe:} pre-training (PT) $\rightarrow$ mid-training (MT) $\rightarrow$ instruct supervised fine-tuning (SFT) $\rightarrow$ quantization-aware training (QAT) with INT4 precision.}
\label{fig:recipe}
\end{figure}

With the MobileMoE architectures derived in Section~\ref{sec:finding_optimal}, we scale them with a four-stage recipe (Figure~\ref{fig:recipe}): pre-training $\rightarrow$ mid-training $\rightarrow$ instruct SFT $\rightarrow$ INT4 QAT, whose design choices are dictated by on-device constraints (Eq.~\eqref{eq:opt-objective}, \eqref{eq:memory}) and the challenges of training MoE models at sub-billion active parameters.

\begin{figure}[!t]
\centering
\includegraphics[width=\linewidth]{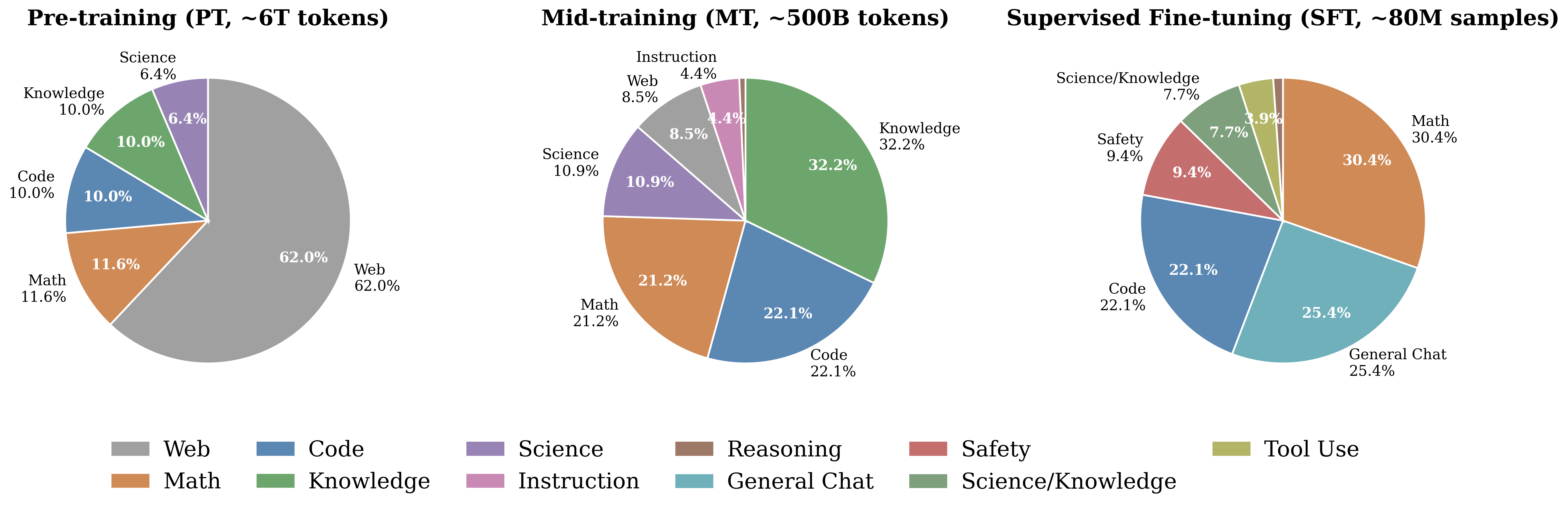}
\caption{\textbf{MobileMoE training data mixtures across stages.} Domain composition of pre-training (PT, $\sim$6T tokens), mid-training (MT, $\sim$500B tokens), and supervised fine-tuning (SFT, $>$80M samples), with consistent colors per domain across stages. Across stages, the mixture transitions from web-heavy coverage in PT toward domain-specific data (knowledge, code, math) in MT and SFT. Full per-domain datasets are listed in Tables~\ref{tab:pt_data}, \ref{tab:mt_data}, and \ref{tab:sft_data}.}
\label{fig:data-mixture}
\end{figure}

\textbf{Pre-training.} We pre-train all three \textbf{MobileMoE-S/M/L} models at context length 2,048, instantiating $\mathrm{MoE}$ with $E=8, g=8, s=\checkmark$ on the base architecture of Figure~\ref{fig:moe-arch}, using tied input-output embeddings, RoPE $\theta{=}500{,}000$, and Llama-3 tokenizer with 128K vocabulary~\cite{llama3}.
All models are trained on ${\sim}6$T tokens from open-licensed data, including a web-heavy mixture (${>}60\%$) to provide broad linguistic coverage for general language modeling, and diverse domain coverage for math, code, knowledge, and science to encourage MoE expert specialization across heterogeneous data and tasks~\cite{shazeer2017outrageously}  (Figure~\ref{fig:data-mixture}, left; full sources in Table~\ref{tab:pt_data}). Although our 6T token budget is smaller than those of recent small LLMs (e.g., 9T for Llama~3.2 1B~\cite{llama3}, 11T for SmolLM2~\cite{smollm2}), MobileMoE remains both token- and compute-efficient: all three model scales continue to improve through 6T training (Figure~\ref{fig:data-scaling}), yet use far fewer training FLOPs to match or surpass the accuracy of those dense LLMs (detailed in Section~\ref{sec:comparison}). To ensure MoE training stability and efficiency, we employ the following techniques.

\textit{MoE training stability.} To ensure stable routing and balanced expert utilization throughout training, we use auxiliary-loss-free balancing~\cite{wang2024auxiliary} (with bias update rate $\lambda_{lb}=10^{-3}$), which adjusts expert biases based on token load imbalance without backpropagating a loss, combined with router z-loss regularization~\cite{zoph2022st} ($\lambda_z=10^{-4}$) to stabilize router logits. We adopt sigmoid gating with per-token top-$k$ normalization, which scores each expert independently rather than forcing competition as in softmax gating, producing smoother routing score distributions. All router computations are performed in FP32 precision for numerical stability.

\textit{MoE training efficiency.} With fine-grained experts, each expert's FFN is much smaller than a standard MoE FFN (e.g., 60 routed FFNs of $768{\times}384$ in MobileMoE-S vs.\ 8 FFNs of $4096{\times}14336$ in Mixtral 8x7B~\cite{mixtral}, $\sim 200{\times}$ smaller), making na\"ive per-expert computation inefficient. We address this with grouped MLP, which batches all experts into a single fused grouped matrix multiplication (GMM) kernel, replacing small sequential GEMMs with one efficient batched operation. To support this efficient batched operation during pre-training, we adopt drop-and-pad token dispatching (capacity factor 1.5), assigning each expert a fixed-size token buffer to ensure uniform buffer sizes for the batched kernel. We also apply expert parallelism ($\mathrm{EP}=4$), allowing each GPU to hold and compute $60/4=15$ routed experts per GPU in MobileMoE for memory efficiency. 

\textbf{Mid-training.} Following pre-training, we perform mid-training to extend context length from 2,048 to 8,192 while shifting the data distribution toward higher-quality, domain-specific data (math, code, knowledge, science) (Figure~\ref{fig:data-mixture}, middle; full sources in Table~\ref{tab:mt_data}). This domain-concentrated mixture further sharpens MoE expert specialization, allowing routed experts to develop deeper expertise on these domains. With 8K context, MobileMoE's compact KV cache configuration ($n_{kv}=4$, $d_h=64$) keeps its KV cache well within the on-device memory budget (Eq.~\eqref{eq:memory}). We train on $\sim$500B tokens ($\sim$8\% of pre-training budget) with linear learning rate annealing~\cite{llama3,hu2024minicpm}, gradually converging the model on the curated data to produce \textbf{MobileMoE-Base} with strengthened downstream capabilities.

\textbf{Supervised fine-tuning (SFT).} We fine-tune MobileMoE-Base on an open-licensed dataset mixture of $\sim$80M samples spanning diverse domains (e.g., math, code, instruction-following, science, QA) at 8K context length with sequence packing. The data mix is composed of public datasets (Figure~\ref{fig:data-mixture}, right; full sources listed in Table~\ref{tab:sft_data}), where each dataset is sampled in proportion to its size. To ensure expert load balance, we continue to apply the same MoE training stability techniques from pre-training. However, we switch to dropless token dispatching, as the drop-and-pad scheme would discard tokens from structured instruction-response pairs, distorting the learning signal and degrading SFT quality. This stage produces \textbf{MobileMoE-SFT} to unlock comprehensive downstream capabilities.

\textbf{Quantization-aware training (QAT).} To fit MobileMoE within on-device memory budgets (e.g., $\leq$3--5\,GB), we apply INT4 QAT~\cite{jacob2018quantization} to MobileMoE-SFT, following a similar recipe to recent on-device LLMs (e.g., MobileLLM-Pro~\cite{mbllmpro}, which shows that direct post-training quantization (PTQ) can degrade quality). Specifically for MoE, we keep the router in FP32 precision throughout QAT to preserve routing stability under quantized gradients, which adds negligible memory cost ($\sim 0.5\%$ overhead). Concretely, all linear weights (attention, MoE FFNs, and embeddings) are quantized with symmetric group-wise INT4 (group size 32), and activations are dynamically quantized to INT8, with router weights kept in FP32. Formally, the linear weights are quantized via:
\begin{equation}
    \tilde{\mathbf{W}}_g = s_g \cdot \text{clamp}\!\left(\left\lfloor \frac{\mathbf{W}_g}{s_g} \right\rceil, q_{\min}, q_{\max}\right), \quad s_g = \frac{2\,\max(|\mathbf{W}_g|)}{2^b - 1}
    \label{eq:qat}
\end{equation}
where $\mathbf{W}_g$ denotes a contiguous group of weights sharing the same scale factor $s_g$, with group size $g=32$, and $q_{\min}=-2^{b-1}$, $q_{\max}=2^{b-1}-1$ define the INT4 quantization range with $b=4$. Initialized from the SFT checkpoint, we apply QAT on the SFT data with standard cross-entropy loss; this stage produces \textbf{MobileMoE-QAT} with model weight footprints of ${\mathcal{M}_{\text{weight}}}=0.68/1.48/2.75\,$GB on S/M/L (Eq.~\eqref{eq:memory}), all fitting within recent smartphone DRAM budgets for on-device deployment (detailed in Section~\ref{sec:ondevice}).

\section{Experiments}
\label{sec:experiment}

\subsection{Experimental setup}
\label{sec:setup}

\begin{table}[!t]
\centering
\caption{\textbf{MobileMoE training hyperparameters across all stages:} pre-training (PT), mid-training (MT), instruct supervised fine-tuning (SFT), and quantization-aware training (QAT).}
\label{tab:training}
\footnotesize
\begin{tabular}{lcccc}
\toprule
\textbf{Hyperparameter} & \textbf{Pre-training} & \textbf{Mid-training} & \textbf{SFT} & \textbf{QAT} \\
\midrule
Context length & 2,048 & 8,192 & 8,192 & 8,192 \\
Total tokens & $\sim$6T & $\sim$500B & $\sim$126B & $\sim$21B \\
Global batch size & 2,048 / 3,072$^\dagger$ & 512 / 768$^\dagger$ & 256 & 256 \\
Tokens per step & 4.2M / 6.3M$^\dagger$ & 4.2M / 6.3M$^\dagger$ & 2.1M & 2.1M \\
Peak learning rate & $4 \times 10^{-4}$ & $4 \times 10^{-5}$ & $4 \times 10^{-6}$ & $4 \times 10^{-6}$ \\
LR schedule & Cosine & Linear & Cosine & Cosine \\
LR min ratio & 0.1 & 0.1 & 0.0 & 0.0 \\
Warmup steps & 8,000 & 50 & 3,000 & 500 \\
Token dispatch & drop-and-pad & drop-and-pad & dropless & dropless \\
Hardware (H100 nodes) & 16--32 & 16--32 & 4--8 & 4--8 \\
Wall-clock time & 3--4 weeks & $\sim$2 days & $\sim$2--3 days & $\sim$2--3 days \\
\midrule
Optimizer & \multicolumn{4}{c}{AdamW ($\beta_1=0.9$, $\beta_2=0.95$, $\epsilon=10^{-15}$)} \\
Weight decay & \multicolumn{4}{c}{0.1} \\
Gradient clipping & \multicolumn{4}{c}{1.0} \\
Precision & \multicolumn{4}{c}{BF16 (model weights); FP32 (router, optimizer states, gradients)} \\
Sequence packing & \multicolumn{4}{c}{Yes} \\
\midrule
\multicolumn{5}{l}{\textit{QAT-specific}} \\
Weight quantization & --- & --- & --- & INT4 (group size 32) \\
Activation quantization & --- & --- & --- & INT8 \\
Embedding quantization & --- & --- & --- & INT4 \\
\bottomrule
\multicolumn{5}{l}{\footnotesize $^\dagger$ First value applies to MobileMoE-S and -M; second value to MobileMoE-L.}
\end{tabular}
\end{table}

\textbf{Training setup.} All four training stages of MobileMoE share the same optimizer setup. We use AdamW ($\beta_1=0.9$, $\beta_2=0.95$, $\epsilon=10^{-15}$) with weight decay 0.1 and gradient clipping at 1.0. Pre-training, mid-training, and SFT use BF16 model weights, with optimizer states, gradients, and MoE router weights kept in FP32 for numerical stability; QAT additionally applies INT4 weight and INT8 activation quantization (Section~\ref{sec:training_recipes}). Learning rate (LR) schedules are stage-specific: pre-training uses a cosine schedule with peak LR $4{\times}10^{-4}$; mid-training uses linear decay from $4{\times}10^{-5}$; SFT and QAT use cosine decay from $4{\times}10^{-6}$. The peak LR decreases by $10\times$ from pre-training through SFT, with QAT inheriting the SFT LR. Training is performed on 8-GPU NVIDIA H100 (96\,GB) nodes. Full per-stage hyperparameters are summarized in Table~\ref{tab:training}.

\textbf{Evaluation protocols.} We evaluate MobileMoE on a comprehensive suite of benchmarks across two capability tiers. The foundational tier covers 14 widely-used benchmarks spanning five core competencies -- commonsense (HellaSwag~\cite{hellaswag}, PIQA~\cite{piqa}, SIQA~\cite{siqa}, WinoGrande~\cite{winogrande}), knowledge (MMLU~\cite{mmlu}, NaturalQuestions~\cite{naturalquestions}, TriviaQA~\cite{triviaqa}), science (ARC-C/E~\cite{arc}, OpenBookQA~\cite{openbookqa}), reading (BoolQ~\cite{boolq}, DROP~\cite{drop}), and reasoning (BBH~\cite{bbh}, GSM8K~\cite{gsm8k}). The advanced tier comprises 8 benchmarks probing frontier capabilities that small LLMs typically struggle with, including math (MATH-500~\cite{math500}, GSM-Plus~\cite{gsmplus}), code (HumanEval pass@1~\cite{humaneval}, MBPP~\cite{mbpp}), instruction following (IFEval~\cite{ifeval}, IFBench~\cite{ifbench}), and knowledge \& reasoning (MMLU-Pro~\cite{mmlupro}, GPQA~\cite{gpqa}). Detailed evaluation configurations on all benchmarks are given in Appendix~\ref{app:eval}.

\textbf{Baseline models.} We compare MobileMoE against state-of-the-art on-device and efficient LLMs with comparable parameter scales: Gemma~3 (270M, 1B)~\cite{gemma3report2025}, SmolLM2 (360M, 1.7B)~\cite{smollm2}, MobileLLM-Pro (1.1B)~\cite{mbllmpro}, Llama~3.2 (1B)~\cite{llama3}, OLMo~2 (1B)~\cite{olmo2}, Qwen3.5 (0.8B, 2B)~\cite{qwen35}, Granite~3.0 (1B, 3B)~\cite{granite2024granite}, Jet-Nemotron (2B, 4B)~\cite{gu2026jet}, and the state-of-the-art MoE model OLMoE-1B-7B (1.3B active, 6.9B total)~\cite{olmoe}. To ensure fair comparison, we re-evaluate all baseline models under identical settings with greedy decoding for reproducibility. Model sources are listed in Appendix~\ref{app:baselines}.

\subsection{Experimental results}
\label{sec:comparison}

\subsubsection{Results of Scaling MobileMoE}

\begin{figure}[!t]
\centering
\includegraphics[width=\linewidth]{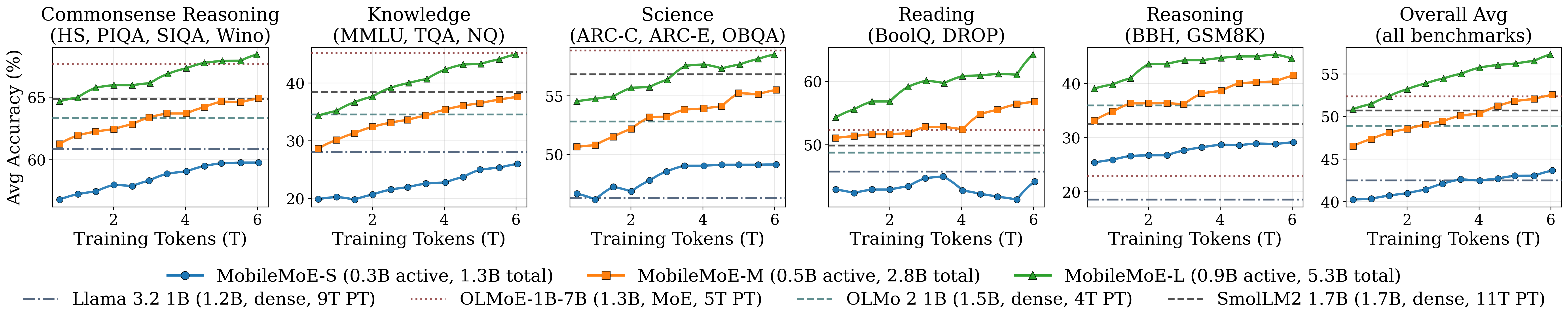}
\caption{\textbf{MobileMoE pre-training trajectories.} Average benchmark accuracy of MobileMoE-S/M/L across five core competencies plus overall average, plotted against training tokens $D$ up to 6T. Horizontal lines: publicly released pre-trained models with reported pre-training token budgets (i.e., Llama~3.2 1B: 9T, with distillation from Llama~3.1 8B; OLMoE-1B-7B: 5T; OLMo~2-1B: 4T; SmolLM2-1.7B: 11T). Evaluation setups are given in Table~\ref{tab:eval-config}.}
\label{fig:data-scaling}
\end{figure}

\textbf{Data scaling results of MobileMoE.} We empirically validate the MobileMoE models derived from on-device scaling laws (Section~\ref{sec:finding_optimal}) at full-scale pre-training. Figure~\ref{fig:data-scaling} plots benchmark performance when scaling training data $D$ up to 6T tokens, and compares against existing baselines with reported training token budgets. We have the following observations. \textbf{(1)} \textit{MobileMoE benefits from continued data scaling, especially on knowledge.} All three MobileMoE scales improve monotonically through 6T on overall average (MobileMoE-S: 40$\rightarrow$44; MobileMoE-M: 46$\rightarrow$53; MobileMoE-L: 51$\rightarrow$57), with the steepest slope on knowledge (MobileMoE-L: 34$\rightarrow$45 on MMLU+TQA+NQ), reflecting MoE's larger total capacity to absorb more knowledge per token than dense models with similar active parameters. \textbf{(2)} \textit{MobileMoE-L exhibits superior token efficiency over both dense and MoE baselines.} It surpasses Llama~3.2 1B (9T, distilled, Avg 42) already at $\sim$0.5T tokens, SmolLM2-1.7B (11T, Avg 51) at $\sim$1T tokens, and OLMoE-1B-7B (5T, MoE, Avg 52) at $\sim$2T tokens on overall average. \textbf{(3)} \textit{MobileMoE achieves superior parameter efficiency over existing MoE.} By 6T pre-training, MobileMoE-L (Avg 57, 922M active) outperforms OLMoE-1B-7B base model (Avg 52, 1.3B active) with 30\% fewer active parameters.

\begin{figure}[!t]
\centering
\includegraphics[width=\linewidth]{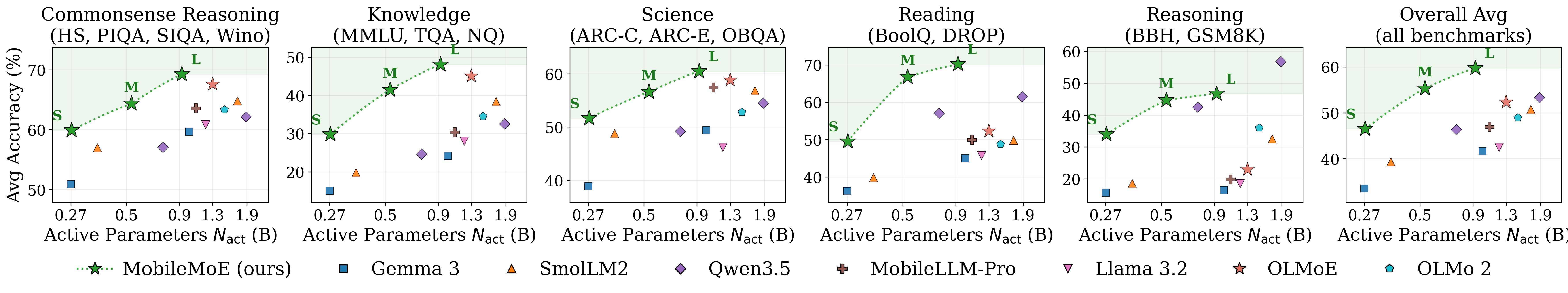}
\caption{\textbf{MobileMoE-Base model comparison.} Average benchmark accuracy of MobileMoE-S/M/L after mid-training across five core competencies plus overall average, vs.\ publicly released base models (Gemma~3, SmolLM2, Qwen3.5, MobileLLM-Pro, Llama~3.2, OLMoE, OLMo~2), plotted against active parameters $N_{\text{act}}$. Green stars: MobileMoE-S/M/L; dashed line: MobileMoE Pareto curve. Full table results are given in Table~\ref{tab:base}.}
\label{fig:model-scaling}
\end{figure}

\textbf{Model scaling results of MobileMoE.} We validate the MobileMoE family at full model scales (S/M/L) after pre-training and mid-training, comparing \textbf{MobileMoE-Base} against publicly released base models. Figure~\ref{fig:model-scaling} plots benchmark performance against active parameters $N_{\text{act}}$ for MobileMoE-Base vs.\ existing base models, with detailed per-benchmark numbers in Table~\ref{tab:base}. We have the following observations. \textbf{(1)} \textit{MobileMoE-Base establishes a new Pareto frontier in the sub-billion active-parameter regime.} Overall, MobileMoE-L (922M active, Avg 60) surpasses OLMoE-1B-7B base model (1.3B active, Avg 52) by +8 with 30\% fewer active parameters, and outperforms the dense baselines at matched or smaller $N_{\text{act}}$. \textbf{(2)} \textit{MobileMoE improves monotonically across S/M/L scales.} The smooth, monotonic MobileMoE curve across S/M/L (Overall Avg: 47$\rightarrow$55$\rightarrow$60) validates that the MoE architecture derived in Section~\ref{sec:finding_optimal} generalizes consistently from scaling-law ablation regime ($\leq$500B tokens) to the full pre- and mid-training, confirming the predictive power of our on-device MoE scaling law. \textbf{(3)} \textit{MobileMoE outperforms dense baselines across 4 out of 5 capability domains.} At matched $N_{\text{act}}$, the largest gains are on knowledge and reading -- reflecting MoE's expanded capacity for knowledge-intensive tasks; the exception is reasoning, where Qwen3.5-2B remains stronger at larger scale with $\sim$2$\times$ more active parameters than MobileMoE-L.

\begin{table}[!t]
\centering
\caption{\textbf{Base model comparison on foundational benchmarks.} $N_{\text{act}}/N_{\text{total}}$: active / total parameters (including embeddings). Superscripts on benchmarks denote few-shot count (0-shot is omitted). \textbf{MobileMoE-S/M/L} refer to 272M/528M/922M active and 1.3B/2.8B/5.3B total params. HS: HellaSwag, Wino: WinoGrande, NQ: NaturalQuestions, TQA: TriviaQA, ARC-C/E: ARC-Challenge/Easy, OBQA: OpenBookQA, BBH: BIG-Bench Hard. All values are benchmark accuracy (\%). All models are evaluated using \href{https://github.com/EleutherAI/lm-evaluation-harness}{lm-eval} with the per-task setup in Table~\ref{tab:eval-config}; full evaluation details in Appendix~\ref{app:eval-foundational}.}
\label{tab:base}
\setlength{\tabcolsep}{2pt}
\resizebox{\linewidth}{!}{%
\begin{tabular}{@{}lc cccc ccc ccc cc cc c@{}}
\toprule
& & \multicolumn{4}{c}{\textbf{Commonsense Reasoning}} & \multicolumn{3}{c}{\textbf{Knowledge}} & \multicolumn{3}{c}{\textbf{Science}} & \multicolumn{2}{c}{\textbf{Reading}} & \multicolumn{2}{c}{\textbf{Reasoning}} & \\
\cmidrule(lr){3-6} \cmidrule(lr){7-9} \cmidrule(lr){10-12} \cmidrule(lr){13-14} \cmidrule(lr){15-16}
\textbf{Model} & $N_{\text{act}}/N_{\text{total}}$ & HS & PIQA & SIQA & Wino & MMLU$^5$ & NQ$^5$ & TQA$^5$ & ARC-C$^{25}$ & ARC-E & OBQA & BoolQ & DROP$^3$ & BBH$^3$ & GSM8K$^8$ & \textbf{Avg} \\
\midrule
Gemma~3 270M & 270M & 41.4 & 68.3 & 40.2 & 53.7 & 26.7 & 4.1 & 14.3 & 29.4 & 56.8 & 30.4 & 58.3 & 14.2 & 29.5 & 1.8 & 33.5 \\
SmolLM2 360M & 362M & 56.5 & 71.7 & 40.7 & \textbf{59.0} & 25.2 & 7.4 & 26.8 & 40.5 & 68.1 & \textbf{37.6} & \textbf{61.8} & 17.9 & 31.7 & 5.3 & 39.3 \\
\textbf{MobileMoE-S} & \textbf{272M/1.3B} & \textbf{58.9} & \textbf{75.4} & \textbf{46.8} & 58.6 & \textbf{43.7} & \textbf{12.6} & \textbf{33.2} & \textbf{46.5} & \textbf{73.9} & 34.6 & 60.2 & \textbf{39.0} & \textbf{31.8} & \textbf{36.2} & \textbf{46.5} \\
\midrule
Qwen3.5 0.8B & 749M & 54.9 & 71.3 & 42.1 & 59.9 & 48.2 & 6.2 & 19.5 & 44.0 & 67.6 & 36.0 & 74.6 & 39.6 & \textbf{40.9} & 44.1 & 46.4 \\
\textbf{MobileMoE-M} & \textbf{528M/2.8B} & \textbf{68.3} & \textbf{77.5} & \textbf{50.2} & \textbf{61.6} & \textbf{54.7} & \textbf{20.9} & \textbf{49.0} & \textbf{51.0} & \textbf{79.4} & \textbf{39.4} & \textbf{75.1} & \textbf{58.5} & 37.7 & \textbf{51.6} & \textbf{55.4} \\
\midrule
Gemma~3 1B & 1.0B & 62.2 & 74.9 & 42.9 & 58.8 & 26.2 & 10.7 & 35.7 & 39.3 & 72.2 & 36.8 & 66.6 & 23.4 & 30.5 & 2.3 & 41.6 \\
MobileLLM-Pro & 1.1B & 66.2 & 76.6 & 48.4 & 63.2 & 32.3 & 15.6 & 43.2 & 52.5 & 76.6 & 43.2 & \textbf{77.5} & 22.5 & 33.0 & 6.6 & 47.0 \\
Llama~3.2 1B & 1.2B & 64.2 & 75.1 & 42.8 & 61.3 & 31.4 & 12.0 & 40.7 & 40.3 & 61.8 & 36.6 & 63.6 & 27.9 & 29.8 & 7.3 & 42.5 \\
OLMo~2 1B & 1.5B & 68.4 & 75.9 & 44.0 & 65.0 & 42.4 & 14.1 & 47.1 & 45.2 & 73.4 & 39.8 & 62.9 & 34.6 & 33.3 & 38.6 & 48.9 \\
SmolLM2 1.7B & 1.7B & 71.4 & 77.6 & 44.2 & 66.1 & 50.2 & 15.4 & 49.6 & 53.3 & 73.4 & 43.8 & 72.4 & 27.3 & 34.0 & 31.0 & 50.7 \\
Qwen3.5 2B & 1.9B & 65.9 & 74.7 & 43.5 & 64.6 & 54.1 & 10.9 & 32.6 & 54.3 & 71.4 & 37.8 & 69.4 & 53.6 & \textbf{48.2} & \textbf{65.3} & 53.3 \\
OLMoE-1B-7B & 1.3B/6.9B & \textbf{77.0} & \textbf{80.5} & 43.9 & \textbf{69.1} & 52.6 & 20.6 & \textbf{62.3} & 55.0 & 76.6 & \textbf{45.0} & 74.8 & 29.8 & 33.5 & 12.3 & 52.4 \\
\textbf{MobileMoE-L} & \textbf{922M/5.3B} & 74.6 & 80.0 & \textbf{54.3} & 68.2 & \textbf{59.6} & \textbf{26.7} & 58.1 & \textbf{57.0} & \textbf{81.7} & 42.8 & 75.7 & \textbf{64.7} & 37.8 & 55.7 & \textbf{59.8} \\
\bottomrule
\end{tabular}
}%
\end{table}

\subsubsection{Results of MobileMoE-Base and MobileMoE-SFT}

\textbf{Results of MobileMoE-Base.} Table~\ref{tab:base} compares MobileMoE-S/M/L against base-model baselines across the 14 foundational benchmarks. We have the following observations.

\textbf{(1)} \textit{MobileMoE-Base achieves state-of-the-art performance among sub-2B active-parameter base models.} MobileMoE-S (272M active, Avg 46.5) surpasses existing models at similar scale: Gemma~3 270M (33.5) by +13.0 and SmolLM2 360M (39.3) by +7.2. MobileMoE-M (528M active, Avg 55.4) outperforms all baselines up to 1.9B active parameters, including Qwen3.5 2B (53.3) by +2.1, OLMoE-1B-7B (1.3B active, 52.4) by +3.0, and SmolLM2 1.7B (50.7) by +4.7. MobileMoE-L (922M active, Avg 59.8) extends this lead, outperforming OLMoE-1B-7B by +7.4 at 30\% fewer active parameters and 23\% fewer total parameters (5.3B vs.\ 6.9B). Notably, MobileMoE-L Base already exceeds the instruct-tuned OLMoE-1B-7B (SFT Avg 55.6, Table~\ref{tab:sft}) by +4.2, confirming that MobileMoE already excels at pre-training and mid-training.

\textbf{(2)} \textit{MobileMoE-Base matches larger dense models at 2-4$\times$ fewer active parameters, with gains concentrated on knowledge and comprehension.} MobileMoE-S (272M, Avg 46.5) approaches Qwen3.5 0.8B (46.4) using $2.8\times$ fewer active parameters. MobileMoE-M (528M, Avg 55.4) surpasses Qwen3.5 2B (53.3) using $3.6\times$ fewer active parameters. MobileMoE-L (922M, Avg 59.8) outperforms Qwen3.5 2B by +6.5 using $\sim$ 2$\times$ fewer active parameters. When we compare both MoE models (MobileMoE-L and OLMoE-1B-7B) against the dense $1$--$2$B LLMs (Qwen3.5 2B, SmolLM2 1.7B, Llama~3.2 1B, Gemma 3 1B, MobileLLM-Pro, OLMo 2 1B), the MoE advantage is strongest on knowledge (MMLU, NQ, TQA) and reading (BoolQ, DROP), which suggests the MoE architecture with larger total parameter capacity is best exploited on tasks demanding stored factual knowledge and comprehension.

\begin{table}[!t]
\centering
\caption{\textbf{Instruct model comparison on foundational benchmarks.} Same conventions are adopted following Table~\ref{tab:base}.}
\label{tab:sft}
\setlength{\tabcolsep}{2pt}
\resizebox{\linewidth}{!}{%
\begin{tabular}{@{}lc cccc ccc ccc cc cc c@{}}
\toprule
& & \multicolumn{4}{c}{\textbf{Commonsense Reasoning}} & \multicolumn{3}{c}{\textbf{Knowledge}} & \multicolumn{3}{c}{\textbf{Science}} & \multicolumn{2}{c}{\textbf{Reading}} & \multicolumn{2}{c}{\textbf{Reasoning}} & \\
\cmidrule(lr){3-6} \cmidrule(lr){7-9} \cmidrule(lr){10-12} \cmidrule(lr){13-14} \cmidrule(lr){15-16}
\textbf{Model} & $N_{\text{act}}/N_{\text{total}}$ & HS & PIQA & SIQA & Wino & MMLU$^5$ & NQ$^5$ & TQA$^5$ & ARC-C$^{25}$ & ARC-E & OBQA & BoolQ & DROP$^3$ & BBH$^3$ & GSM8K$^8$ & \textbf{Avg} \\
\midrule
Gemma~3 270M & 270M & 39.4 & 67.1 & 39.6 & 53.0 & 26.5 & 2.8 & 9.1 & 27.7 & 50.5 & 35.0 & 56.1 & 11.0 & 31.8 & 5.8 & 32.5 \\
SmolLM2 360M & 362M & 56.9 & 71.6 & 40.6 & 57.4 & 25.9 & 6.4 & 20.4 & 38.8 & 49.1 & 36.2 & 42.5 & 15.2 & 30.5 & 10.0 & 35.8 \\
Granite-3.0-1B & 400M/1.3B & \textbf{64.5} & \textbf{75.2} & \textbf{49.7} & \textbf{61.9} & 31.9 & 8.4 & 29.3 & 40.9 & 64.4 & \textbf{39.0} & 69.8 & 16.2 & 32.0 & 30.0 & 43.8 \\
\textbf{MobileMoE-S} & \textbf{272M/1.3B} & 56.5 & 74.9 & 42.2 & 58.4 & \textbf{42.6} & \textbf{10.8} & \textbf{30.2} & \textbf{43.1} & \textbf{73.4} & 32.4 & \textbf{72.3} & \textbf{32.3} & \textbf{32.2} & \textbf{52.2} & \textbf{46.7} \\
\midrule
Qwen3.5 0.8B & 749M & 49.7 & 69.4 & 38.8 & 57.6 & 50.2 & 3.3 & 16.1 & 41.8 & 61.4 & 30.8 & 62.5 & 33.3 & 37.8 & 45.7 & 42.7 \\
Granite-3.0-3B & 800M/3.3B & \textbf{71.8} & \textbf{77.6} & \textbf{54.3} & \textbf{66.1} & 50.2 & 16.7 & 41.9 & 50.6 & 74.7 & \textbf{40.8} & \textbf{77.5} & 35.6 & 37.4 & 52.2 & 53.4 \\
\textbf{MobileMoE-M} & \textbf{528M/2.8B} & 66.6 & \textbf{77.6} & 49.0 & 63.0 & \textbf{53.9} & \textbf{17.5} & \textbf{46.6} & \textbf{52.5} & \textbf{79.9} & 38.2 & 76.7 & \textbf{46.6} & \textbf{39.0} & \textbf{67.5} & \textbf{55.3} \\
\midrule
Gemma~3 1B & 1.0B & 57.8 & 72.3 & 42.0 & 59.0 & 40.0 & 7.6 & 23.4 & 40.1 & 63.1 & 38.4 & 75.8 & 22.0 & 35.8 & 38.9 & 44.0 \\
MobileLLM-Pro & 1.1B & 52.2 & 73.2 & 42.7 & 51.7 & 38.7 & 8.9 & 19.9 & 35.8 & 52.7 & 29.6 & 68.8 & 25.5 & 29.1 & 31.8 & 40.0 \\
Llama~3.2 1B & 1.2B & 61.7 & 74.8 & 43.1 & 61.5 & 46.1 & 14.4 & 38.4 & 42.1 & 63.7 & 37.8 & 71.6 & 21.5 & 33.9 & 46.0 & 46.9 \\
OLMoE-1B-7B & 1.3B/6.9B & \textbf{78.8} & \textbf{79.7} & 50.8 & \textbf{68.7} & 52.7 & 17.2 & 54.1 & 57.6 & 75.9 & \textbf{46.8} & \textbf{81.1} & 29.3 & 37.1 & 49.1 & 55.6 \\
OLMo~2 1B & 1.5B & 67.3 & 75.2 & 46.1 & 63.5 & 42.9 & 12.4 & 37.6 & 45.1 & 69.8 & 42.2 & 71.0 & 31.2 & 35.0 & 46.9 & 49.0 \\
SmolLM2 1.7B & 1.7B & 71.7 & 76.2 & 44.6 & 68.4 & 49.4 & 14.2 & 46.0 & 53.4 & 62.9 & 45.8 & 68.5 & 24.8 & 35.3 & 46.1 & 50.5 \\
Qwen3.5 2B & 1.9B & 62.2 & 72.8 & 41.0 & 63.0 & 57.4 & 8.8 & 28.1 & 53.2 & 66.0 & 35.2 & 71.7 & 44.7 & 45.2 & 61.3 & 50.8 \\
Jet-Nemotron-2B & 2.0B & 64.3 & 73.4 & 38.9 & 64.0 & 54.4 & 15.1 & 38.4 & 54.6 & 57.4 & 34.6 & 77.5 & 54.4 & 42.2 & \textbf{78.3} & 53.4 \\
Jet-Nemotron-4B & 4.0B & 70.7 & 75.0 & 41.3 & 68.0 & 57.8 & 18.6 & 48.8 & 56.7 & 58.3 & 38.0 & 75.3 & \textbf{55.3} & \textbf{46.2} & 77.6 & 56.3 \\
\textbf{MobileMoE-L} & \textbf{922M/5.3B} & 73.0 & 78.9 & \textbf{53.4} & 66.1 & \textbf{60.1} & \textbf{22.4} & \textbf{54.9} & \textbf{57.9} & \textbf{81.9} & 43.2 & \textbf{81.1} & 50.1 & 40.1 & 77.6 & \textbf{60.1} \\
\bottomrule
\end{tabular}
}%
\end{table}

\textbf{Results of MobileMoE-SFT.} Table~\ref{tab:sft} compares MobileMoE-S/M/L against state-of-the-art baselines across the same benchmarks on diverse domains, under identical evaluation protocols. We report average scores across benchmarks as a holistic measure of capability, and summarize our findings as follows.

\textbf{(1)} \textit{MobileMoE-SFT achieves state-of-the-art performance among sub-2B active-parameter instruct models, preserving the performance advantages of MobileMoE-Base.} MobileMoE-S (272M active, Avg 46.7) surpasses existing instruct models at similar scale: Gemma~3 270M (32.5) by +14.2 and SmolLM2 360M (35.8) by +10.9. MobileMoE-M (528M active, Avg 55.3) outperforms all dense baselines up to 1.9B active parameters, including Qwen3.5 2B (50.8) by +4.5, SmolLM2 1.7B (50.5) by +4.8, and OLMo~2 1B (49.0) by +6.3. MobileMoE-L (922M active, Avg 60.1) further extends this lead, surpassing the larger MoE baseline OLMoE-1B-7B (1.3B active, 55.6) by +4.5 despite using 30\% fewer active parameters and 23\% fewer total parameters (5.3B vs.\ 6.9B). These margins match or exceed the gaps prior to SFT (Table~\ref{tab:base}), confirming our SFT recipe preserves MobileMoE's architectural advantages.

\textbf{(2)} \textit{MobileMoE-SFT matches dense instruct models using 2--4$\times$ fewer active parameters, with the same parameter-efficiency Pareto frontier as MobileMoE-Base.} MobileMoE-S (272M, Avg 46.7) surpasses Qwen3.5 0.8B (42.7) by +4.0 using $2.8\times$ fewer active parameters, and matches Llama~3.2 1B (46.9) with $4.4\times$ fewer active parameters. MobileMoE-M (528M, Avg 55.3) outperforms Qwen3.5 2B (50.8) by +4.5 using $3.6\times$ fewer active parameters. MobileMoE-L (922M, Avg 60.1) exceeds Qwen3.5 2B by +9.3 using $\sim$2$\times$ fewer active parameters, with especially strong gains on knowledge, science, and math. These results demonstrate that MobileMoE-SFT delivers comparable or superior quality to larger dense instruct models with 2--4$\times$ fewer inference FLOPs, translating to lower latency and reduced power consumption for efficient on-device deployment.

\begin{table}[!t]
\centering
\caption{\textbf{Instruct model comparison on advanced benchmarks.} \textbf{Avg} is within-capability mean; \textbf{Overall} is mean across all. All models are evaluated in non-thinking mode with the same setup, using \href{https://github.com/EleutherAI/lm-evaluation-harness}{lm-eval} and official packages \href{https://github.com/TIGER-AI-Lab/MMLU-Pro}{TIGER-AI-Lab/MMLU-Pro}, \href{https://github.com/allenai/IFBench}{allenai/IFBench} (detailed setup in Table~\ref{tab:eval-config-instruct}; more analysis in Appendix~\ref{app:eval-variants}).}
\label{tab:sft-appendix}
\setlength{\tabcolsep}{5pt}
\resizebox{\linewidth}{!}{%
\begin{tabular}{@{}lc!{\vrule}ccc!{\vrule}ccc!{\vrule}ccc!{\vrule}ccc!{\vrule}c@{}}
\toprule
& & \multicolumn{3}{c!{\vrule}}{\textbf{Math}} & \multicolumn{3}{c!{\vrule}}{\textbf{Code}} & \multicolumn{3}{c!{\vrule}}{\textbf{Instruction Following}} & \multicolumn{3}{c!{\vrule}}{\textbf{Knowledge \& Reasoning}} & \\
\cmidrule(lr){3-5} \cmidrule(lr){6-8} \cmidrule(lr){9-11} \cmidrule(lr){12-14}
\textbf{Model} & $N_{\text{act}}/N_{\text{total}}$ & MATH500$^4$ & GSM+$^5$ & \textbf{Avg} & HumanEval & MBPP$^3$ & \textbf{Avg} & IFEval & IFBench & \textbf{Avg} & MMLU-Pro$^5$ & GPQA & \textbf{Avg} & \textbf{Overall} \\
\midrule
Gemma~3 270M & 270M & \phantom{0}7.2 & \phantom{0}4.3 & \phantom{0}5.7 & 12.8 & \phantom{0}9.8 & 11.3 & 31.2 & 11.2 & 21.2 & 11.3 & 25.8 & 18.6 & 14.2 \\
SmolLM2 360M & 362M & \phantom{0}3.8 & \phantom{0}4.6 & \phantom{0}4.2 & \phantom{0}0.0 & 22.8 & 11.4 & 40.2 & \textbf{19.1} & 29.7 & 12.0 & 25.8 & 18.9 & 16.0 \\
Granite-3.0-1B & 400M/1.3B & 11.2 & 16.0 & 13.6 & 20.7 & \textbf{28.8} & 24.8 & 37.2 & 11.2 & 24.2 & 13.9 & 25.8 & 19.8 & 20.6 \\
\textbf{MobileMoE-S} & \textbf{272M/1.3B} & \textbf{21.0} & \textbf{28.9} & \textbf{24.9} & \textbf{44.5} & 27.8 & \textbf{36.2} & \textbf{54.1} & 13.8 & \textbf{33.9} & \textbf{18.2} & \textbf{27.8} & \textbf{23.0} & \textbf{29.5} \\
\midrule
Qwen3.5 0.8B & 749M & 19.6 & 26.5 & 23.1 & 31.1 & 25.4 & 28.3 & 59.9 & 19.8 & 39.8 & 24.0 & \textbf{26.3} & 25.1 & 29.1 \\
Granite-3.0-3B & 800M/3.3B & 20.6 & 31.8 & 26.2 & 15.2 & 41.2 & 28.2 & 45.0 & 16.1 & 30.5 & 23.0 & 24.8 & 23.9 & 27.2 \\
\textbf{MobileMoE-M} & \textbf{528M/2.8B} & \textbf{27.2} & \textbf{42.3} & \textbf{34.7} & \textbf{60.4} & \textbf{43.0} & \textbf{51.7} & \textbf{60.8} & \textbf{20.9} & \textbf{40.8} & \textbf{28.3} & 24.8 & \textbf{26.5} & \textbf{38.4} \\
\midrule
Gemma~3 1B & 1.0B & 27.6 & 25.2 & 26.4 & 42.1 & 40.6 & 41.3 & 63.7 & 17.5 & 40.6 & 16.1 & 25.3 & 20.7 & 32.3 \\
MobileLLM-Pro & 1.1B & \phantom{0}8.8 & 17.0 & 12.9 & 59.8 & 44.2 & 52.0 & 63.1 & 17.0 & 40.1 & 10.9 & 23.2 & 17.1 & 30.5 \\
Llama~3.2 1B & 1.2B & 19.6 & 27.8 & 23.7 & 36.6 & 37.4 & 37.0 & 58.7 & 17.4 & 38.0 & 20.8 & 28.8 & 24.8 & 30.9 \\
OLMoE-1B-7B & 1.3B/6.9B & \phantom{0}8.4 & 28.1 & 18.2 & 36.0 & 30.2 & 33.1 & 48.1 & 16.6 & 32.4 & 19.5 & 24.2 & 21.9 & 26.4 \\
OLMo~2 1B & 1.5B & 10.2 & 25.1 & 17.7 & 29.3 & 14.8 & 22.0 & 53.5 & 14.5 & 34.0 & 16.0 & 29.8 & 22.9 & 24.1 \\
SmolLM2 1.7B & 1.7B & 15.4 & 26.6 & 21.0 & \phantom{0}1.2 & 34.6 & 17.9 & 54.7 & 16.5 & 35.6 & 19.8 & 29.8 & 24.8 & 24.8 \\
Qwen3.5 2B & 1.9B & 31.0 & 42.4 & 36.7 & 50.0 & 41.2 & 45.6 & \textbf{73.3} & \textbf{30.3} & \textbf{51.8} & 38.8 & 34.3 & 36.6 & 42.7 \\
Jet-Nemotron-2B & 2.0B & \textbf{43.2} & 53.0 & 48.1 & 59.1 & 54.0 & 56.6 & 36.5 & 10.0 & 23.3 & 39.0 & \textbf{37.9} & 38.5 & 41.6 \\
Jet-Nemotron-4B & 4.0B & 43.0 & \textbf{54.1} & \textbf{48.6} & 64.6 & \textbf{57.4} & \textbf{61.0} & 44.1 & 11.2 & 27.7 & \textbf{44.4} & 36.9 & \textbf{40.7} & \textbf{44.5} \\
\textbf{MobileMoE-L} & \textbf{922M/5.3B} & 32.2 & 50.2 & 41.2 & \textbf{65.2} & 52.4 & 58.8 & 67.3 & 20.1 & 43.7 & 34.0 & 33.8 & 33.9 & 44.4 \\
\bottomrule
\end{tabular}
}%
\end{table}

\textbf{Results of MobileMoE-SFT on additional capabilities.} Table~\ref{tab:sft-appendix} probes four advanced capabilities where sub-billion dense baselines often collapse (e.g., SmolLM2-360M HumanEval$=0.0$, Gemma~3 270M MATH500$=7.2$). We have these observations: \textbf{(1)} \textit{MobileMoE shows strong code and math performance at sub-billion active-parameter scales.} On code, MobileMoE-S/M/L (Avg 36.2/51.7/58.8) outperform scale-matched baselines substantially (e.g., MobileMoE-L vs.\ Qwen3.5 2B +13.2, vs.\ OLMoE-1B-7B +25.7). On math, MobileMoE-L (Avg 41.2) leads Qwen3.5 2B (Avg 36.7) by +4.5 and OLMoE-1B-7B (Avg 18.2) by +23.0; MobileMoE-S (Avg 24.9) scores substantially higher than Gemma~3 270M (Avg 5.7) and SmolLM2-360M (Avg 4.2). \textbf{(2)} \textit{MobileMoE remains competitive on Instruction Following (IF) and Knowledge \& Reasoning (K\&R).} MobileMoE-L outperforms OLMoE-1B-7B on both capabilities and both Jet-Nemotron variants on IF; it trails Qwen3.5 2B on IF and Jet-Nemotron-4B on K\&R while using substantially fewer active parameters. The Qwen3.5 2B advantage likely reflects its more advanced post-training recipe (e.g., distillation, thinking-enabled reasoning). Overall, the MoE benefits of larger total capacity and routed expert specialization yield clear strengths on code and math, while the gap with Qwen3.5 2B on instruction following and knowledge \& reasoning motivates enriching our training recipe in future work, e.g., adding distillation and thinking-enabled post-training as in Qwen3/3.5~\cite{qwen3,qwen35}.

\begin{figure}[!t]
\centering
\includegraphics[width=\linewidth]{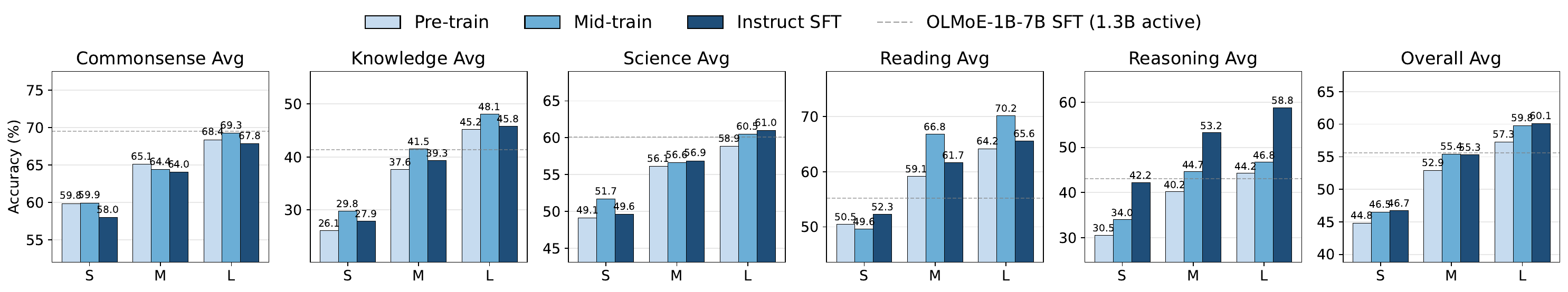}
\caption{\textbf{MobileMoE capability progression across training stages.} Benchmark accuracy (\%) for MobileMoE-S/M/L after each of the three main training stages: pre-training (PT, light), mid-training (MT, medium), and instruction supervised fine-tuning (SFT, dark). Gray dashed line: instruct-tuned OLMoE-1B-7B (1.3B active) as a scale-matched MoE reference baseline. Each per-competence panel (Commonsense, Knowledge, Science, Reading, Reasoning) reports the average over its constituent benchmarks; the overall panel reports the average over all 14 benchmarks. Full per-benchmark numerical results are in Table~\ref{tab:stages}.}
\label{fig:stages}
\end{figure}

\begin{table}[!t]
\centering
\caption{\textbf{MobileMoE training stage progression with scale-matched reference baselines.} Detailed benchmark results for MobileMoE-S/M/L across pre-training (PT), mid-training (MT), and instruction SFT stages, together with Base and SFT results for baselines of similar scales (SmolLM2 360M, Qwen3.5 0.8B, OLMoE-1B-7B). Benchmarks and few-shot setup follow Table~\ref{tab:base}.}
\label{tab:stages}
\setlength{\tabcolsep}{2pt}
\resizebox{\linewidth}{!}{%
\begin{tabular}{@{}llc cccc ccc ccc cc cc c@{}}
\toprule
& & & \multicolumn{4}{c}{\textbf{Commonsense Reasoning}} & \multicolumn{3}{c}{\textbf{Knowledge}} & \multicolumn{3}{c}{\textbf{Science}} & \multicolumn{2}{c}{\textbf{Reading}} & \multicolumn{2}{c}{\textbf{Reasoning}} & \\
\cmidrule(lr){4-7} \cmidrule(lr){8-10} \cmidrule(lr){11-13} \cmidrule(lr){14-15} \cmidrule(lr){16-17}
\textbf{Model} & \textbf{Stage} & $N_{\text{act}}/N_{\text{total}}$ & HS & PIQA & SIQA & Wino & MMLU$^5$ & NQ$^5$ & TQA$^5$ & ARC-C$^{25}$ & ARC-E & OBQA & BoolQ & DROP$^3$ & BBH$^3$ & GSM8K$^8$ & \textbf{Avg} \\
\midrule
\multirow{2}{*}{SmolLM2 360M} & Base & \multirow{2}{*}{362M} & 56.5 & 71.7 & 40.7 & 59.0 & 25.2 & 7.4 & 26.8 & 40.5 & 68.1 & 37.6 & 61.8 & 17.9 & 31.7 & 5.3 & 39.3 \\
& SFT  &  & 56.9 & 71.6 & 40.6 & 57.4 & 25.9 & 6.4 & 20.4 & 38.8 & 49.1 & 36.2 & 42.5 & 15.2 & 30.5 & 10.0 & 35.8 \\
\cmidrule(l){1-17}
\multirow{3}{*}{\textbf{MobileMoE-S}} & PT  & \multirow{3}{*}{\textbf{272M/1.3B}} & 60.5 & 73.1 & 46.8 & 59.0 & 33.5 & 11.9 & 33.0 & 45.0 & 67.5 & 34.8 & 61.9 & 39.0 & 31.7 & 29.4 & 44.8 \\
& MT  &  & 58.9 & 75.4 & 46.8 & 58.6 & 43.7 & 12.6 & 33.2 & 46.5 & 73.9 & 34.6 & 60.2 & 39.0 & 31.8 & 36.2 & 46.5 \\
& SFT &  & 56.5 & 74.9 & 42.2 & 58.4 & 42.6 & 10.8 & 30.2 & 43.1 & 73.4 & 32.4 & 72.3 & 32.3 & 32.2 & 52.2 & 46.7 \\
\midrule[\heavyrulewidth]
\multirow{2}{*}{Qwen3.5 0.8B} & Base & \multirow{2}{*}{749M} & 54.9 & 71.3 & 42.1 & 59.9 & 48.2 & 6.2 & 19.5 & 44.0 & 67.6 & 36.0 & 74.6 & 39.6 & 40.9 & 44.1 & 46.4 \\
& SFT  &  & 49.7 & 69.4 & 38.8 & 57.6 & 50.2 & 3.3 & 16.1 & 41.8 & 61.4 & 30.8 & 62.5 & 33.3 & 37.8 & 45.7 & 42.7 \\
\cmidrule(l){1-17}
\multirow{3}{*}{\textbf{MobileMoE-M}} & PT  & \multirow{3}{*}{\textbf{528M/2.8B}} & 68.7 & 77.0 & 51.4 & 63.3 & 49.5 & 16.5 & 46.8 & 52.4 & 74.9 & 41.0 & 71.1 & 47.2 & 34.0 & 46.4 & 52.9 \\
& MT  &  & 68.3 & 77.5 & 50.2 & 61.6 & 54.7 & 20.9 & 49.0 & 51.0 & 79.4 & 39.4 & 75.1 & 58.5 & 37.7 & 51.6 & 55.4 \\
& SFT &  & 66.6 & 77.6 & 49.0 & 63.0 & 53.9 & 17.5 & 46.6 & 52.5 & 79.9 & 38.2 & 76.7 & 46.6 & 39.0 & 67.5 & 55.3 \\
\midrule[\heavyrulewidth]
\multirow{2}{*}{OLMoE-1B-7B} & Base & \multirow{2}{*}{1.3B/6.9B} & 77.0 & 80.5 & 43.9 & 69.1 & 52.6 & 20.6 & 62.3 & 55.0 & 76.6 & 45.0 & 74.8 & 29.8 & 33.5 & 12.3 & 52.4 \\
& SFT  &  & 78.8 & 79.7 & 50.8 & 68.7 & 52.7 & 17.2 & 54.1 & 57.6 & 75.9 & 46.8 & 81.1 & 29.3 & 37.1 & 49.1 & 55.6 \\
\cmidrule(l){1-17}
\multirow{3}{*}{\textbf{MobileMoE-L}} & PT  & \multirow{3}{*}{\textbf{922M/5.3B}} & 74.3 & 79.4 & 52.5 & 67.3 & 55.5 & 22.6 & 57.6 & 57.0 & 75.6 & 44.0 & 74.2 & 54.2 & 36.3 & 52.2 & 57.3 \\
& MT  &  & 74.6 & 80.0 & 54.3 & 68.2 & 59.6 & 26.7 & 58.1 & 57.0 & 81.7 & 42.8 & 75.7 & 64.7 & 37.8 & 55.7 & 59.8 \\
& SFT &  & 73.0 & 78.9 & 53.4 & 66.1 & 60.1 & 22.4 & 54.9 & 57.9 & 81.9 & 43.2 & 81.1 & 50.1 & 40.1 & 77.6 & 60.1 \\
\bottomrule
\end{tabular}
}%
\end{table}

\textbf{Comparison of capabilities across training stages.} Figure~\ref{fig:stages} traces how MobileMoE-S/M/L evolve across three main training stages -- pre-training (PT), mid-training (MT), and instruction supervised fine-tuning (SFT). We also compare to the Base and SFT checkpoints of three baselines at comparable scales (SmolLM2 360M for MoE-S, Qwen3.5 0.8B for MoE-M, OLMoE-1B-7B for MoE-L) as references on all 14 foundational benchmarks (Table~\ref{tab:stages}). We highlight four observations.

\textbf{(1)} \textit{Pre-training dominates commonsense reasoning.} HellaSwag, PIQA, SIQA, and WinoGrande saturate by the end of PT and remain within $\pm$2 points through MT and SFT across all three MobileMoE scales, indicating that broad linguistic priors are largely acquired from the 6T-token pre-training mixture and do not substantially improve in later mid-training or SFT.

\textbf{(2)} \textit{Mid-training drives the largest knowledge and reading gains.} Upweighting curated knowledge, code, math, and long-document sources yields the largest PT$\to$MT jumps on MMLU (MobileMoE-S: $33.5{\to}43.7$, $+10.2$; M: $49.5{\to}54.7$, $+5.2$; L: $55.5{\to}59.6$, $+4.1$) and DROP (M: $47.2{\to}58.5$, $+11.3$; L: $54.2{\to}64.7$, $+10.5$), confirming that mid-training is the primary mechanism for strengthening factual recall and long-context comprehension.

\textbf{(3)} \textit{Instruction SFT unlocks reasoning.} GSM8K exhibits the most dramatic MT$\to$SFT jumps (L: $55.7{\to}77.6$, $+21.9$; M: $51.6{\to}67.5$, $+15.9$; S: $36.2{\to}52.2$, $+16.0$), alongside consistent BoolQ and BBH gains, demonstrating that instruction tuning is essential for eliciting multi-step chain-of-thought reasoning. Overall foundational accuracy improves $+2$ to $+3$ points from PT to MT across all three MobileMoE scales, while SFT preserves foundational accuracy and unlocks the math/reasoning gains in Table~\ref{tab:sft-appendix}.

\textbf{(4)} \textit{MobileMoE-L's advantage over OLMoE-1B-7B is established at PT and compounds through SFT.} As shown in Table~\ref{tab:base}, MobileMoE-L PT already exceeds OLMoE-1B-7B Instruct on the average benchmark; after SFT, MobileMoE-L further surpasses OLMoE-1B-7B on the majority of benchmarks (dashed reference line in Figure~\ref{fig:stages}), confirming that MobileMoE's advantage establishes during pre-training and retains through subsequent stages.

\subsection{On-device deployment of MobileMoE}
\label{sec:ondevice}

We deploy MobileMoE to flagship smartphones (\emph{Samsung Galaxy~S25} with Snapdragon~8~Elite, \emph{iPhone~16~Pro} with Apple~A18~Pro) via INT4 QAT (Section~\ref{sec:training_recipes}) and custom MoE inference implemented on \href{https://github.com/pytorch/executorch}{ExecuTorch}. We benchmark against MobileLLM-Pro~\cite{mbllmpro} -- a state-of-the-art on-device LLM purpose-built for mobile deployment, providing a strong baseline at comparable parameter count. In the following section, we evaluate MobileMoE on two key fronts for on-device deployment: (1) benchmark performance with INT4 weight precision, and (2) on-device runtime profiling across different mobile devices and processors.

\begin{table}[!t]
\centering
\caption{\textbf{Quantized model comparison with quantization-aware training (INT4 QAT).} All models are quantized to 4-bit weights via QAT. \textbf{Mem}: INT4 model weight memory in GB. Baseline QAT checkpoint: MobileLLM-Pro~(\href{https://huggingface.co/facebook/MobileLLM-Pro-base-int4-accelerator}{\texttt{facebook/MobileLLM-Pro-base-int4-accelerator}}). Benchmark abbreviations follow Table~\ref{tab:base}.}
\label{tab:qat}
\setlength{\tabcolsep}{2pt}
\resizebox{\linewidth}{!}{%
\begin{tabular}{@{}lc cccc ccc ccc cc cc c@{}}
\toprule
& & \multicolumn{4}{c}{\textbf{Commonsense Reasoning}} & \multicolumn{3}{c}{\textbf{Knowledge}} & \multicolumn{3}{c}{\textbf{Science}} & \multicolumn{2}{c}{\textbf{Reading}} & \multicolumn{2}{c}{\textbf{Reasoning}} & \\
\cmidrule(lr){3-6} \cmidrule(lr){7-9} \cmidrule(lr){10-12} \cmidrule(lr){13-14} \cmidrule(lr){15-16}
\textbf{Model} & Mem (GB) & HS & PIQA & SIQA & Wino & MMLU$^5$ & NQ$^5$ & TQA$^5$ & ARC-C$^{25}$ & ARC-E & OBQA & BoolQ & DROP$^3$ & BBH$^3$ & GSM8K$^8$ & \textbf{Avg} \\
\midrule
MobileLLM-Pro & 0.55 & 64.7 & 75.6 & 47.4 & 62.8 & 30.4 & 13.9 & 39.9 & 51.6 & 75.2 & \textbf{42.8} & 76.8 & 20.5 & 31.4 & 4.1 & 45.5 \\
\textbf{MobileMoE-S} & 0.68 & 53.5 & 73.5 & 45.5 & 55.7 & 39.8 & 8.6 & 25.0 & 43.3 & 69.7 & 33.2 & 70.8 & 24.6 & 31.4 & 40.8 & 44.0 \\
\textbf{MobileMoE-M} & 1.48 & 63.7 & 76.7 & 49.0 & 61.2 & 52.4 & 5.0 & 42.1 & 51.2 & 79.1 & 36.6 & 75.8 & 43.2 & 36.2 & 62.6 & 52.5 \\
\textbf{MobileMoE-L} & 2.75 & \textbf{71.0} & \textbf{79.0} & \textbf{52.9} & \textbf{65.2} & \textbf{57.0} & \textbf{19.4} & \textbf{52.4} & \textbf{55.1} & \textbf{80.3} & 42.4 & \textbf{78.1} & \textbf{44.6} & \textbf{38.3} & \textbf{73.2} & \textbf{57.8} \\
\bottomrule
\end{tabular}
}%
\end{table}

\subsubsection{Results of MobileMoE-QAT}

We apply INT4 QAT (Section~\ref{sec:training_recipes}, Eq.~\eqref{eq:qat}) to all linear layers, yielding INT4 weight memory footprints of 0.55/0.68/1.48/2.75\,GB for MobileLLM-Pro and MobileMoE-S/M/L. Unlike MobileLLM-Pro's publicly released pre-trained QAT model, MobileMoE is quantized on top of its SFT models for direct on-device deployment. Table~\ref{tab:qat} compares the per-benchmark accuracy of MobileMoE-S/M/L against the publicly released QAT baseline MobileLLM-Pro. We highlight our observations.

\textbf{(1)} \textit{QAT preserves nearly all of MobileMoE's BF16 accuracy at $4\times$ weight compression.} Compared against the corresponding BF16 SFT checkpoints (Table~\ref{tab:sft}), MobileMoE's QAT incurs a 2--3 point drop on overall average: MobileMoE-S $46.7{\to}44.0$ ($-2.7$), MobileMoE-M $55.3{\to}52.5$ ($-2.8$), MobileMoE-L $60.1{\to}57.8$ ($-2.3$). This degradation is small compared to the $4\times$ weight-memory reduction from BF16 to INT4, indicating that MoE routing and expert computation remain numerically stable under 4-bit quantization and that our QAT recipe (symmetric group-wise INT4 weights + FP32 router, Section~\ref{sec:training_recipes}) effectively recovers the BF16 capability.

\textbf{(2)} \textit{MobileMoE-QAT compresses to on-device memory budgets while remaining competitive with QAT baselines.} After INT4 QAT, MobileMoE-S/M/L have model weight memory ($\mathcal{M}_{\text{weight}}$) of $0.68$/$1.48$/$2.75$\,GB, with the theoretical memory proxy $\mathcal{M}$ (Eq.~\eqref{eq:memory}, weights $+$ INT8 KV cache at $8\text{k}$ context) of $0.76$/$1.58$/$2.88$\,GB -- all comfortably within modern mobile DRAM budgets ($\leq 5$\,GB). At comparable INT4 weight memory, MobileMoE-S ($\mathcal{M}_{\text{weight}}=0.68$\,GB, Avg $44.0$) matches MobileLLM-Pro ($\mathcal{M}_{\text{weight}}=0.55$\,GB, $45.5$) within $1.5$ points on overall average while substantially raising knowledge results (MMLU $+9.4$). The INT4 QAT MobileMoE-L ($\mathcal{M}_{\text{weight}}=2.75$\,GB, Avg $57.8$) already exceeds the BF16 SFT OLMoE-1B-7B (Avg $55.6$, $\sim$13.8\,GB BF16) at $\sim$$5\times$ smaller memory footprint.

\subsubsection{On-device runtime profiling}

\textbf{On-device MoE inference.} Existing mobile CPU inference backends such as XNNPACK provide highly optimized INT4 dense matmul kernels but lack a fused MoE feed-forward operator. We therefore implement a custom MoE operator in ExecuTorch guided by two principles: (1) \emph{convert sparse expert dispatch into dense grouped GEMMs} -- we first reorder tokens (via a counting sort on their assigned expert IDs) so that all tokens routed to the same expert sit contiguously in memory, letting each expert process its slice of tokens as a single dense batched matmul through torchao's INT4 GEMM kernel; and (2) \emph{fuse every sub-operation inside each MoE FFN layer into a single op} -- top-$k$ expert selection (over router logits), token dispatch, per-expert gate- and up-projections (fused into one GEMM per expert), SwiGLU activation, down-projection, and weighted-scatter unpermute share one op call, amortizing kernel-launch and activation-quantization overhead. Attention and embedding layers continue to use the XNNPACK INT4 dense path, so MobileLLM-Pro runs end-to-end on XNNPACK while MobileMoE routes its MoE FFN blocks through our custom op.

\textbf{Runtime profiling setup.} We set up runtime profiling on two flagship smartphones: \emph{Samsung Galaxy~S25} (Snapdragon~8~Elite, using 4 CPU threads) and \emph{iPhone~16~Pro} (Apple A18 Pro, using 2 CPU threads). We use the same ExecuTorch CPU backend with XNNPACK and batch size 1. Weights are INT4 symmetric with group size 32 (torchao INT4 packing); activations are INT8 dynamic per-row. We sweep input sequence lengths across two generation regimes: (1) short-context generation, e.g., on-device chat, with input $\in \{256, 512, 1024\}$ tokens and output $=128$ tokens; and (2) long-context generation, e.g., on-device summarization, with input $\in \{2048, 4096, 8192\}$ tokens and output $=1024$ tokens. The expert utilization ratio of MoE is input-dependent: the portions of activated experts vary according to the input token types (Appendix~\ref{app:quantitative}). Therefore, rather than following the standard LLM runtime profiling that feeds dummy prompts of fixed sequence lengths (e.g., repeating the same words or using random tokens), we use real prompts across different domains (knowledge, code, math), and run inference 3 times per prompt to compute the average runtime benchmark.

\begin{table}[!t]
\centering
\caption{\textbf{On-device runtime latency}: comparing MobileMoE-S/M/L against the dense MobileLLM-Pro on two flagship smartphones -- Samsung Galaxy~S25 (Snapdragon~8~Elite, 4 CPU threads) and iPhone~16~Pro (Apple A18 Pro, 2 CPU threads), both via ExecuTorch+XNNPACK backend with INT4 weights and INT8 dynamic activations. \textbf{Prefill TTFT} (s, $\downarrow$): time-to-first-token. \textbf{Decode Rate} (tok/s, $\uparrow$): generation throughput. Both averaged over multiple runs on real prompts (code, knowledge, math). \textbf{Mem} (GB): static INT4 weight memory; \textbf{Avg} (\%, $\uparrow$): mean accuracy over 14 benchmarks after QAT (Table \ref{tab:qat}).  Best results per column are in \textbf{bold}.}
\label{tab:latency}
\setlength{\tabcolsep}{2.0pt}
\resizebox{\linewidth}{!}{%
\begin{tabular}{@{}l!{\vrule}cc!{\vrule}cccccc!{\vrule}cccccc!{\vrule}cccccc!{\vrule}cccccc@{}}
\toprule
& & & \multicolumn{12}{c!{\vrule}}{\textbf{Samsung Galaxy~S25}} & \multicolumn{12}{c}{\textbf{iPhone~16~Pro}} \\
\cmidrule(lr){4-15} \cmidrule(lr){16-27}
& \textbf{Mem} & \textbf{Avg} & \multicolumn{6}{c!{\vrule}}{\textbf{Prefill TTFT (s)} $\downarrow$} & \multicolumn{6}{c!{\vrule}}{\textbf{Decode Rate (tok/s)} $\uparrow$} & \multicolumn{6}{c!{\vrule}}{\textbf{Prefill TTFT (s)} $\downarrow$} & \multicolumn{6}{c}{\textbf{Decode Rate (tok/s)} $\uparrow$} \\
\textbf{Model} & (GB) & (\%) & 256 & 512 & 1k & 2k & 4k & 8k & 256 & 512 & 1k & 2k & 4k & 8k & 256 & 512 & 1k & 2k & 4k & 8k & 256 & 512 & 1k & 2k & 4k & 8k \\
\midrule
MobileLLM-Pro & \textbf{0.55} & 45.5 & 0.78 & 1.73 & 4.26 & 14.21 & 35.62 & 105.21 & 61.3 & 56.5 & 45.8 & 27.6 & 18.6 & 10.6 & 1.29 & 2.90 & 6.03 & 14.13 & 39.08 & 122.42 & 61.0 & 55.1 & 48.5 & 32.4 & 17.9 & 10.6 \\
\textbf{MobileMoE-S} & 0.68 & 44.0 & \textbf{0.44} & \textbf{0.90} & \textbf{2.01} & \textbf{7.82} & \textbf{16.36} & \textbf{50.33} & \textbf{138.1} & \textbf{130.0} & \textbf{112.0} & \textbf{61.7} & \textbf{48.9} & \textbf{25.8} & \textbf{0.46} & \textbf{1.02} & \textbf{2.14} & \textbf{5.23} & \textbf{13.86} & \textbf{39.91} & \textbf{204.6} & \textbf{180.5} & \textbf{148.4} & \textbf{90.2} & \textbf{54.2} & \textbf{32.2} \\
\textbf{MobileMoE-M} & 1.48 & 52.5 & 0.84 & 1.79 & 4.32 & 16.68 & 35.02 & 88.25 & 83.6 & 77.1 & 64.6 & 34.5 & 26.2 & 15.4 & 1.01 & 2.23 & 4.71 & 10.82 & 31.43 & 83.54 & 106.8 & 95.3 & 76.2 & 51.0 & 27.9 & 17.2 \\
\textbf{MobileMoE-L} & 2.75 & \textbf{57.8} & 1.53 & 4.09 & 8.99 & 25.85 & 55.12 & 162.45 & 53.4 & 43.0 & 36.6 & 22.9 & 17.3 & 8.9 & 1.86 & 4.32 & 8.63 & 21.92 & 51.98 & 141.31 & 59.4 & 51.9 & 43.3 & 23.5 & 17.1 & 10.8 \\
\bottomrule
\end{tabular}
}%
\end{table}

\textbf{Results of on-device runtime profiling on latency.} Table~\ref{tab:latency} reports the full runtime sweep over $\{256, 512, 1\text{k}, 2\text{k}, 4\text{k}, 8\text{k}\}$ sequence lengths on Samsung Galaxy~S25 and iPhone 16 Pro, comparing MobileMoE-S/M/L to MobileLLM-Pro, deployed with INT4 quantization and ExecuTorch+XNNPACK backend. We highlight these findings.

\textbf{(1)} \textit{MobileMoE-S achieves a Pareto win at every context length on both smartphones:} at comparable INT4 weights ($0.68$ vs.\ $0.55$\,GB) and similar accuracy (Avg $44.0$ vs.\ $45.5$) to MobileLLM-Pro, on \emph{Samsung S25} MobileMoE-S is $1.8$--$2.2\times$ faster at prefill and $2.2$--$2.6\times$ faster at decode. On \emph{iPhone~16~Pro}, the latency speedup widens further to $2.7$--$3.1\times$ at prefill and $2.8$--$3.4\times$ at decode. These results demonstrate that the architectural advantage of MobileMoE-S generalizes across mobile devices with different silicon (Qualcomm vs.\ Apple), attributed to its much smaller compute FLOPs and per-token memory bandwidth.

\textbf{(2)} \textit{MobileMoE-M/L delivers substantially higher accuracy ($\mathit{+7.0}$/$\mathit{+12.3}$ Avg over MobileLLM-Pro) with comparable or modest runtime cost:} on \emph{Samsung S25}, MobileMoE-M (Avg $52.5$ vs.\ $45.5$, with ${+7.0}$ boost) achieves prefill parity at short context and $1.0$--$1.2\times$ at $\geq 4\text{k}$ (overall $0.9$--$1.2\times$), with a steady $1.3$--$1.5\times$ decode speedup; MobileMoE-L (Avg $57.8$ vs.\ $45.5$, with ${+12.3}$ boost), the highest-accuracy variant, delivers this large accuracy gain at a moderate runtime drop ($0.4$--$0.7\times$ prefill, $0.8$--$0.9\times$ decode) that shrinks with context. On \emph{iPhone~16~Pro}, runtime speedup ratios over the dense baseline strengthen further: MobileMoE-M outperforms MobileLLM-Pro at $1.2$--$1.5\times$ prefill and $1.6$--$1.8\times$ decode, while MobileMoE-L narrows its drop to $0.6$--$0.9\times$ prefill and $0.7$--$1.0\times$ decode (parity at $8\text{k}$ decode) with substantial performance gain. These results demonstrate that MobileMoE-M/L's substantial accuracy gains at competitive runtime hold consistently across mobile devices, establishing a new on-device Pareto frontier beyond MobileLLM-Pro at higher accuracy.

\begin{table}[!t]
\centering
\caption{\textbf{MoE runtime speedup across mobile devices and processors (CPU, GPU)}: Samsung Galaxy~S25 (Snapdragon~8~Elite CPU, XNNPACK), iPhone~16~Pro (Apple A18 Pro CPU, XNNPACK), and iPhone~16~Pro (Apple A18 Pro Metal GPU, MLX). \textbf{Prefill TTFT} (s, $\downarrow$): time-to-first-token. \textbf{Decode Rate} (tok/s, $\uparrow$): generation throughput. Both averaged over multiple runs. \textbf{Mem} (GB): static INT4 weight memory; \textbf{Avg} (\%, $\uparrow$): mean accuracy over 14 benchmarks after QAT (Table \ref{tab:qat}). The \textbf{MoE Speedup} row reports MobileMoE-S vs.\ MobileLLM-Pro per metric per context.}
\label{tab:speedup}
\setlength{\tabcolsep}{5pt}
\resizebox{\linewidth}{!}{%
\begin{tabular}{@{}l!{\vrule}cc!{\vrule}ccc!{\vrule}ccc!{\vrule}ccc!{\vrule}ccc!{\vrule}ccc!{\vrule}ccc@{}}
\toprule
& & & \multicolumn{6}{c!{\vrule}}{\textbf{Samsung S25 (Snapdragon CPU, XNNPACK)}} & \multicolumn{6}{c!{\vrule}}{\textbf{iPhone 16 Pro (Apple CPU, XNNPACK)}} & \multicolumn{6}{c}{\textbf{iPhone 16 Pro (Apple GPU, MLX)}} \\
\cmidrule(lr){4-9} \cmidrule(lr){10-15} \cmidrule(lr){16-21}
& \textbf{Mem} & \textbf{Avg} & \multicolumn{3}{c!{\vrule}}{\textbf{Prefill TTFT (s)} $\downarrow$} & \multicolumn{3}{c!{\vrule}}{\textbf{Decode (tok/s)} $\uparrow$} & \multicolumn{3}{c!{\vrule}}{\textbf{Prefill TTFT (s)} $\downarrow$} & \multicolumn{3}{c!{\vrule}}{\textbf{Decode (tok/s)} $\uparrow$} & \multicolumn{3}{c!{\vrule}}{\textbf{Prefill TTFT (s)} $\downarrow$} & \multicolumn{3}{c}{\textbf{Decode (tok/s)} $\uparrow$} \\
\textbf{Model} & (GB) & (\%) & 512 & 1k & 2k & 512 & 1k & 2k & 512 & 1k & 2k & 512 & 1k & 2k & 512 & 1k & 2k & 512 & 1k & 2k \\
\midrule
MobileLLM-Pro & \textbf{0.55} & \textbf{45.5 }& 1.73 & 4.26 & 14.21 & 56.5 & 45.8 & 27.6 & 2.90 & 6.03 & 14.13 & 55.1 & 48.5 & 32.4 & 0.58 & 1.20 & 2.49 & 61.8 & 59.2 & 56.3 \\
\textbf{MobileMoE-S} & 0.68 & 44.0 & \textbf{0.90} & \textbf{2.01} & \textbf{7.82} & \textbf{130.0} & \textbf{112.0} & \textbf{61.7} & \textbf{1.02} & \textbf{2.14} & \textbf{5.23} & \textbf{180.5} & \textbf{148.4} & \textbf{90.2} & \textbf{0.16} & \textbf{0.32} & \textbf{0.68} & \textbf{154.3} & \textbf{151.3} & \textbf{141.9} \\
\midrule
\textbf{MoE Speedup} & --- & --- & $1.9\times$ & $2.1\times$ & $1.8\times$ & $2.3\times$ & $2.5\times$ & $2.2\times$ & $2.8\times$ & $2.8\times$ & $2.7\times$ & $3.3\times$ & $3.1\times$ & $2.8\times$ & $3.6\times$ & $3.8\times$ & $3.7\times$ & $2.5\times$ & $2.6\times$ & $2.5\times$ \\
\bottomrule
\end{tabular}
}%
\end{table}

\textbf{Why can MoE achieve better on-device latency?} As shown in Table~\ref{tab:speedup}, MobileMoE-S yields consistent $1.8$--$3.8\times$ prefill and $2.2$--$3.4\times$ decode speedups over the dense MobileLLM-Pro across mobile silicon (Qualcomm vs.\ Apple), processors (CPU vs.\ GPU), and inference backends (XNNPACK vs.\ MLX), at comparable accuracy and INT4 model memory. This advantage stems from MoE's much smaller active-parameter count ($<\frac{1}{3}$ of dense at comparable accuracy), which reduces both per-token compute and memory bandwidth at inference. Prefill is \emph{compute-bound}: per-token FFN matmul scales with active (not total) parameters, so MoE's smaller $N_\text{active}$ shrinks the dominant prefill cost. Decode is \emph{memory-bandwidth-bound}: per-step weight reads from RAM also scale with active parameters, so MoE transfers fewer bytes per token, yielding higher throughput. Critically, our custom MoE kernel turns these theoretical FLOPs and bandwidth savings into measurable on-device speedups, empirically realizing the compute and memory efficiency derived in Section~\ref{sec:finding_optimal}.

\begin{table}[!t]
\centering
\caption{\textbf{On-device peak runtime memory}: comparing MobileMoE-S/M/L against the dense MobileLLM-Pro on Samsung Galaxy~S25 (Snapdragon~8~Elite, 4 CPU threads), via ExecuTorch+XNNPACK backend with INT4 weights and INT8 dynamic activations. \textbf{Peak RSS} (GB, $\downarrow$): maximum runtime RAM during inference, including resident weights, KV cache, transient activations, and runtime overhead, averaged over multiple runs. We report Peak RSS separately under \emph{real prompts} (code, knowledge, math) and \emph{dummy prompts} (repeated tokens). \textbf{Mem} (GB): static INT4 weight memory; \textbf{Avg} (\%, $\uparrow$): mean accuracy over 14 benchmarks after QAT (Table \ref{tab:qat}).  Best results per column are in \textbf{bold}.}
\label{tab:memory}
\setlength{\tabcolsep}{8pt}
\resizebox{0.9\linewidth}{!}{%
\begin{tabular}{@{}l!{\vrule}cc!{\vrule}cccccc!{\vrule}cccccc@{}}
\toprule
& & & \multicolumn{6}{c!{\vrule}}{\textbf{Peak RSS, real prompts (GB)} $\downarrow$} & \multicolumn{6}{c}{\textbf{Peak RSS, dummy prompts (GB)} $\downarrow$} \\
\textbf{Model} & \textbf{Mem (GB)} & \textbf{Avg (\%)} & 256 & 512 & 1k & 2k & 4k & 8k & 256 & 512 & 1k & 2k & 4k & 8k \\
\midrule
MobileLLM-Pro & \textbf{0.55} & 45.5 & \textbf{0.90} & \textbf{0.93} & \textbf{0.98} & \textbf{1.07} & 1.35 & 1.91 & 0.90 & 0.92 & 0.98 & 1.07 & 1.32 & 1.87 \\
\textbf{MobileMoE-S} & 0.68 & 44.0 & 0.93 & 0.97 & 1.02 & 1.10 & \textbf{1.23} & \textbf{1.49} & \textbf{0.63} & \textbf{0.61} & \textbf{0.75} & \textbf{0.85} & \textbf{0.86} & \textbf{1.11} \\
\textbf{MobileMoE-M} & 1.48 & 52.5 & 1.98 & 2.04 & 2.12 & 2.24 & 2.43 & 2.77 & 1.10 & 1.27 & 1.15 & 1.20 & 1.44 & 1.91 \\
\textbf{MobileMoE-L} & 2.75 & \textbf{57.8} & 3.66 & 3.75 & 3.87 & 4.06 & 4.27 & 4.71 & 1.72 & 1.99 & 1.84 & 2.41 & 3.49 & 3.41 \\
\bottomrule
\end{tabular}
}%
\end{table}

\textbf{Results of on-device profiling on peak runtime memory.} Table~\ref{tab:memory} reports Peak RSS on Samsung Galaxy S25 over different sequence lengths on real prompts and dummy prompts, where Peak RSS is the maximum runtime RAM during inference, including resident weights, KV cache, transient activations, and runtime overhead, which is therefore larger than the static INT4 weight memory and can be optimized and reduced via various runtime memory optimizations (e.g., paged KV cache, activation reuse, kernel fusion). We use Peak RSS here as a conservative upper bound on on-device memory usage. We highlight these findings.

\textbf{(1)} \textit{Real prompts are essential for valid MoE memory profiling:} Recall that our runtime profiling is conducted with real prompts (code, knowledge, math) rather than dummy prompts (repeated tokens). On Samsung S25, Peak RSS under real prompts rises to $1.2$--$2.1\times$ that of dummy prompts for MobileMoE-S/M/L, while staying at $\sim1.0\times$ for the dense MobileLLM-Pro. This asymmetry reflects MoE's input-dependent expert routing at runtime: real prompts activate diverse experts and load more expert weights into RAM, whereas dummy prompts trigger a narrow routing pattern that loads fewer experts. Dummy-prompt RSS thus captures only the \emph{lower bound} of MoE memory usage, while real-prompt RSS faithfully reflects actual runtime behavior.

\textbf{(2)} \textit{MobileMoE-S matches dense RAM at short context and saves substantial RAM at long context:} At comparable INT4 weight memory ($0.68$ vs.\ $0.55$\,GB), MobileMoE-S Peak RSS is within $\sim$$5\%$ of MobileLLM-Pro at short context ($\leq 1\text{k}$) and substantially lower at long context -- $9\%$ lower at $4\text{k}$ ($1.23$ vs.\ $1.35$\,GB) and $22\%$ lower at $8\text{k}$ ($1.49$ vs.\ $1.91$\,GB), which is driven by fewer transformer layers (smaller KV cache), narrower $d_\text{model}$ (smaller activation buffers), and the mmap loading of only activated experts into RAM. Notably, despite having slightly more total parameters than the dense baseline, the runtime memory and latency advantages of MobileMoE-S hold at comparable INT4 weight footprint.

\textbf{(3)} \textit{MobileMoE-M/L trade extra RAM for substantially higher accuracy, all fitting within commodity on-device DRAM budgets:} relative to MobileLLM-Pro, MobileMoE-M's Peak RSS overhead falls from $\sim 2.2\times$ at short context to $1.45\times$ at $8\text{k}$, and MobileMoE-L's from $4.1\times$ at $256$ to $2.5\times$ at $8\text{k}$. Even at the largest scale, MobileMoE-L's Peak RSS stays under $5$\,GB at $8\text{k}$ context ($4.71$\,GB) -- comfortably within modern mobile DRAM budgets, confirming that all MobileMoE variants are practical for on-device deployment.

\section{Conclusion}

We presented \textbf{MobileMoE}, a family of on-device MoE language models. Our work develops a generalized on-device MoE scaling law that jointly optimizes architecture under mobile memory and compute constraints, an end-to-end recipe that scales MobileMoE training to establish a new Pareto frontier for on-device LLMs in benchmark performance, and the first efficient on-device MoE deployment on commodity smartphone CPUs, with systematic profiling across CPU and GPU backends, which together demonstrate MoE as a practical path at the edge. Several promising directions can further build on this work. On the MoE training side, distillation, reasoning-oriented post-training, and multimodal extensions could unlock better performance and capabilities. On the MoE runtime side, dynamic routing, model compression (e.g., expert pruning, mixed-precision quantization), and mobile NPU deployment could yield further on-device efficiency. While existing on-device models remain predominantly dense, we show that MoE offers a more efficient alternative. We hope MobileMoE opens new directions for next-generation on-device AI, bringing capable, efficient sparse LLMs to edge devices such as smartphones, wearables, and embodied agents, enabling local intelligence that lowers cloud compute demand while delivering private, low-latency inference.

\section*{Author Contributions}\label{sec:contributions}

Yanbei Chen designed the project scope, developed the MobileMoE model architecture (Sections~\ref{sec:preliminaries},~\ref{sec:scaling-law},~\ref{sec:finding_optimal}), the end-to-end training recipe (Section~\ref{sec:training_recipes}) and data (Appendix~\ref{app:data}), ran all experiments (Sections~\ref{sec:setup},~\ref{sec:comparison}), and wrote the paper. Hanxian Huang set up the evaluation pipelines (Section~\ref{sec:setup}) and contributed to post-training data (Appendix~\ref{app:data}). Ernie Chang curated the pre-training and mid-training data (Appendix~\ref{app:data}). Jacob Szwejbka deployed MobileMoE to iPhone~16~Pro and benchmarked with CPU and GPU backends (Section~\ref{sec:ondevice}). Digant Desai implemented the custom MoE kernel for mobile inference, and deployed MobileMoE to Samsung Galaxy~S25 with runtime benchmarks (Section~\ref{sec:ondevice}). Zechun Liu and Vikas Chandra provided feedback through regular meetings and project-planning discussions. Raghuraman Krishnamoorthi advised on research direction, provided feedback through regular discussions, and supported the project with computing and other resources. All authors contributed to discussions that shaped the scope and direction of this work.

\section*{Acknowledgements}

We thank Wei Wen, Igor Fedorov, Yanghan Wang, Tarek Elgamal, Qi Qian, Kimish Patel, Steven Li, Bilgin Cagatay, Mergen Nachin, Min Guo, Shicong Zhao, Patrick Huber, Rylan Conway, Hakan Boyraz, Amrit Panda, Eugene Gorbatov, and Harshit Khaitan for valuable discussions, help, and feedback on this project.

\bibliographystyle{assets/plainnat}
\bibliography{references}

\clearpage
\newpage
\beginappendix
\renewcommand{\thetable}{\Alph{section}\arabic{table}}
\renewcommand{\thefigure}{\Alph{section}\arabic{figure}}
\counterwithin{table}{section}
\counterwithin{figure}{section}
\setcounter{table}{0}
\setcounter{figure}{0}

\section{Scaling Law Ablation Details}
\label{app:ablations}

This appendix provides the detailed configurations, parametric fitting procedure, and training efficiency analysis for the three scaling-law ablations -- number of experts $E$, expert granularity $g$, and shared expert $s$ -- that underpin the on-device MoE architecture derivation in Section~\ref{sec:finding_optimal}. Table~\ref{tab:ablations} summarizes the sweep range, fixed settings, and resulting finding of each ablation. All ablations use the base architectures in Figure~\ref{fig:moe-arch} and train on up to $\sim$500B tokens.

\begin{table}[!h]
\centering
\caption{\textbf{Scaling law ablation configurations.} All ablations run on 8 nodes (64 NVIDIA H100 96\,GB GPUs) with global batch size 3,072 and sequence length 2,048, which takes 2-10 days to complete each ablation.}
\label{tab:ablations}
\small
\resizebox{\linewidth}{!}{%
\begin{tabular}{lccccc}
\toprule
\textbf{Ablation} & \textbf{Sweep range} & $N_{\text{act}}$ (B/billion) & $D$ (B/billion tokens) & \textbf{Fixed settings} & \textbf{Finding} \\
\midrule
Sparsity ($E$)       & $E \in \{1, 2, 4, 8, 16, 32\}$ & $\{0.3, 0.5, 0.9\}$ & $\{100, 150, 200, \ldots, 500\}$ & $g=1$, $s=\times$    & Finding 1: $E=8$ \\
Granularity ($g$)    & $g \in \{1, 2, 4, 8, 16\}$     & $\{0.3, 0.5, 0.9\}$ & $\{100, 150, 200, \ldots, 500\}$ & $E=8$, $s=\times$    & Finding 2: $g=8$ \\
Shared expert ($s$)  & $s \in \{\checkmark, \times\}$ & $\{0.3, 0.5, 0.9\}$ & $\{100, 150, 200, \ldots, 500\}$ & $E=8$, $g=8$      & Finding 3: $s=\checkmark$ \\
\bottomrule
\end{tabular}
}
\end{table}

\subsection{Parametric Fitting of On-Device MoE Scaling Laws}
\label{app:fitting}

\paragraph{Parametric fitting procedure.} To estimate the coefficients of the on-device MoE scaling law (Eq.~\eqref{eq:generalized-moe}), we adopt a two-stage procedure: (i) \texttt{scipy.optimize.curve\_fit} (nonlinear least-squares with MSE) provides a warm-start initialization, followed by (ii) \texttt{scipy.optimize.minimize} with L-BFGS-B optimization (similar to~\cite{hoffmann2022chinchilla}) using an MSE objective, under bounds that keep the irreducible loss $c > 0$. Each ablation fits on the validation loss across runs that sweep over varying active parameters $N_{\text{act}} \in \{0.3, 0.5, 0.9\}$\,billion and data tokens $D \in \{100, 150, 200, \ldots, 500\}$\,billion tokens, with the transformed expert count $\hat{E}$ following~\cite{clark2022unified}: $\frac{1}{\hat{E}} = \frac{1}{E - 1 + \left(\frac{1}{E_{\text{start}}} - \frac{1}{E_{\text{max}}}\right)^{-1}} + \frac{1}{E_{\text{max}}}$. We set $E_{\text{start}}=1$ (the dense baseline) and $E_{\text{max}}=32$ (the upper bound beyond which total parameters exceed typical mobile DRAM budgets of 5\,GB at INT4 for our sub-billion active-parameter regime: $N_{\text{act}} \in \{0.3, 0.5, 0.9\}$\,B). We also find empirically that using the simplified form $\hat{E}{=}E$ ($E_{\text{start}}{=}1$, $E_{\text{max}}{=}\infty$; RMSE=0.0089) or fitting $E_{\text{start}}, E_{\text{max}}$ as free parameters (RMSE=0.0060) yields the same optimal $E$ under a 5,GB on-device memory budget; our fitting choice anchors the transformation to the on-device constraint while maintaining competitive fit quality (RMSE=0.0076). We obtain the optimal-$E$ findings by interpolation within the swept range.

\begin{table}[!h]
\centering
\caption{\textbf{Fitted scaling-law coefficients across the three ablations.} Columns list the nine coefficients of Eq.~\eqref{eq:generalized-moe}. The $E$-sweep row directly fits all coefficients of Eq.~\eqref{eq:generalized-moe} with $x$ fixed (i.e., $A=A_x$, $\delta=\delta_x$, $\alpha=\alpha_x$, $\gamma=\gamma_x$, $B=B_x$, $\omega=\omega_x$, $\beta=\beta_x$, $\zeta=\zeta_x$, $c=c_x$). The $g$- and $s$-rows fit Eq.~\eqref{eq:generalized-moe} with $\hat{E}$ fixed; their entries under $A, \alpha, B, \beta$ are the Chinchilla composites $\tilde{A}=A_x\hat{E}^{\delta_x}$, $\tilde{\alpha}=\alpha_x+\gamma_x \ln \hat{E}$, $\tilde{B}=B_x\hat{E}^{\omega_x}$, $\tilde{\beta}=\beta_x+\zeta_x \ln \hat{E}$ (marked with $^\ast$), while $\delta, \gamma, \omega, \zeta$ are absorbed (``abs.''). The irreducible loss $c_x$ reflects the entropy floor of the validation data and is architecture-independent; the $g$- and $s$-sweeps therefore regularize $c_x$ toward the $E$-sweep estimate (std.\ err.\ $\pm 0.13$). RMSE: root-mean-square error of the fit on validation loss.}
\label{tab:fit-all}
\small
\setlength{\tabcolsep}{3pt}
\resizebox{\linewidth}{!}{%
\begin{tabular}{@{}llccccccccc|c@{}}
\toprule
\textbf{Sweep} & \textbf{Setting} & $A_x$ & $\delta_x$ & $\alpha_x$ & $\gamma_x$ & $B_x$ & $\omega_x$ & $\beta_x$ & $\zeta_x$ & $c_x$ & \textbf{RMSE} \\
\midrule
\makecell[l]{$E$-sweep\\(fixed $g=1$, $s=\times$)} & joint fit & 0.2388 & 0.0906 & $-0.2833$ & 0.0387 & 0.6019 & 1.0593 & $-0.3210$ & $-0.3684$ & 1.9730 & 0.0076 \\
\midrule
\multirow{5}{*}{\shortstack[l]{$g$-sweep\\(fixed $E=8$, $s=\times$)}}
  & $g=1$  & 0.2747$^\ast$ & abs. & $-0.2265^\ast$ & abs. & 2.4777$^\ast$ & abs. & $-0.8296^\ast$ & abs. & 1.9730 & 0.0037 \\
  & $g=2$  & 0.1466$^\ast$ & abs. & $-0.3243^\ast$ & abs. & 0.3697$^\ast$ & abs. & $-0.2492^\ast$ & abs. & 1.9687 & 0.0031 \\
  & $g=4$  & 0.1823$^\ast$ & abs. & $-0.2880^\ast$ & abs. & 1.1616$^\ast$ & abs. & $-0.6188^\ast$ & abs. & 1.9744 & 0.0036 \\
  & $g=8$  & 0.1670$^\ast$ & abs. & $-0.3006^\ast$ & abs. & 0.5199$^\ast$ & abs. & $-0.3870^\ast$ & abs. & 1.9636 & 0.0029 \\
  & $g=16$ & 0.1442$^\ast$ & abs. & $-0.3377^\ast$ & abs. & 1.1266$^\ast$ & abs. & $-0.6204^\ast$ & abs. & 2.0054 & 0.0034 \\
\midrule
\multirow{2}{*}{\shortstack[l]{$s$-sweep\\(fixed $E=8$, $g=8$)}}
  & $s=\times$          & 0.1670$^\ast$ & abs. & $-0.3006^\ast$ & abs. & 0.5199$^\ast$ & abs. & $-0.3870^\ast$ & abs. & 1.9636 & 0.0029 \\
  & $s=\checkmark$ & 0.1224$^\ast$ & abs. & $-0.3884^\ast$ & abs. & 0.3487$^\ast$ & abs. & $-0.2185^\ast$ & abs. & 1.9636 & 0.0033 \\
\bottomrule
\end{tabular}
}
\end{table}

\paragraph{Scaling law coefficients.} Table~\ref{tab:fit-all} reports the fitted coefficients across all three ablations.
\begin{itemize}[leftmargin=*, nosep]
    \item \textbf{Scaling the number of experts $E$:} the $E$-sweep fits Eq.~\eqref{eq:generalized-moe} with $x$ fixed, jointly capturing scaling across $E \in \{1, 2, 4, 8, 16, 32\}$, $N_{\text{act}}$, and $D$ on $>100$ datapoints to fit 9 coefficients, and parameterizes the scaling-law curves in Figures~\ref{fig:scaling-experts} and~\ref{fig:moe-optimal-flops}(a) underlying Finding~1.
    \item \textbf{Scaling the expert granularity $g$:} the $g$-sweep fits Eq.~\eqref{eq:generalized-moe} with $\hat{E}$ fixed independently for each $g \in \{1, 2, 4, 8, 16\}$ at $E=8$ and $s=\times$ (27 datapoints per fit), parameterizing the compute-optimal scaling curves in Figure~\ref{fig:moe-optimal-flops}(b) underlying Finding~2.
    \item \textbf{Scaling with shared expert $s$:} the $s$-sweep fits Eq.~\eqref{eq:generalized-moe} with $\hat{E}$ fixed at $E=8$, $g=8$, comparing no shared expert ($s=\times$: $64$ routed experts with top-$8$ routing) against with shared expert ($s=\checkmark$: $60$ routed experts plus one always-on shared expert with top-$4$ routing; see Section~\ref{sec:finding_optimal}), with 27 datapoints per fit, parameterizing the compute-optimal curves in Figure~\ref{fig:moe-optimal-flops}(c) underlying Finding~3.
\end{itemize}
Note that the $E$-sweep requires a joint fit given the $E$-dependent exponents ($\delta_x, \gamma_x, \omega_x, \zeta_x$) are only identifiable when $E$ varies; while the $g$- and $s$-sweeps fit independently since $E$ is held fixed (absorbing those exponents into effective constants) and the scaling dynamics on $g$ and $s$ is characterized by the reduced form in Eq.~\eqref{eq:chinchilla}, $\mathcal{L}_{\hat{E}}(N_{\text{act}}, D, x) = \tilde{A}_x N_{\text{act}}^{\tilde{\alpha}_x} + \tilde{B}_x D^{\tilde{\beta}_x} + c_x$, where the architecture choice $x \in \{g, s\}$ modulates only the effective coefficients $(\tilde{A}_x, \tilde{\alpha}_x, \tilde{B}_x, \tilde{\beta}_x, c_x)$, therefore each setting of $x$ can be fitted separately from its own $(N_{\text{act}}, D)$ grid.

\section{Training Data}
\label{app:data}

All training data across pre-training, mid-training, and SFT stages are publicly available under permissive open-source licenses (CC-BY-4.0, Apache~2.0, ODC-BY, MIT, NVIDIA License). Dataset names below are hyperlinked to their source repositories. Figure~\ref{fig:data-mixture} visualizes the data mixture composition across three training stages; per-domain dataset lists are detailed in Tables~\ref{tab:pt_data}, \ref{tab:mt_data}, \ref{tab:sft_data}.

\paragraph{Pre-training data.} Our pre-training data of MobileMoE comes from two sources: (1) the \href{https://huggingface.co/datasets/allenai/dolma3_mix-6T-1025-7B}{Dolma3 mix} from OLMo-3~\cite{olmo3}, which provides quality-stratified Common Crawl, OCR-extracted scientific PDFs~\cite{olmocr}, multi-language code (Stack-Edu), and curated academic math (FineMath~\cite{smollm2}); and (2) a curated data collection from MobileLLM models~\cite{mobilellmr1,mbllmpro}, including quality-filtered web corpora (DCLM~\cite{dclm}, FineWeb-Edu~\cite{finewebedu}), code (StarCoder~\cite{starcoder}, Stack-Edu, Nemotron Code~\cite{nemotrondata}), math (FineMath~\cite{smollm2}, OpenWebMath~\cite{openwebmath}, Algebraic Stack~\cite{proofpile2}, Nemotron Math~\cite{nemotrondata}), science (arXiv, peS2o, Nemotron Science~\cite{nemotrondata}, Natural Reasoning~\cite{naturalreasoning}), and knowledge (FLAN~\cite{flan}, StackExchange, Wiki, Cosmopedia). The combined pre-training data mix is web-heavy (62\%), as web corpora are abundant and diverse text data, providing broad linguistic coverage to maximize general language modeling capacity. The remaining data is widely distributed across diverse domains: math (11.6\%), knowledge (10\%), code (10\%), and science (6.4\%) to build capabilities in reasoning, coding, and factual knowledge during pre-training. This domain diversity is particularly beneficial for MoE models, as exposure to heterogeneous data encourages expert specialization across different token types~\cite{shazeer2017outrageously}. Table~\ref{tab:pt_data} summarizes the domain breakdown.

\begin{table}[!h]
\centering
\caption{\textbf{Pre-training data sources by domain.} Dataset names are linked to their source repositories.}
\label{tab:pt_data}
\footnotesize
\setlength{\tabcolsep}{4pt}
\begin{tabular}{l p{0.72\linewidth} r}
\toprule
\textbf{Domain} & \textbf{Key Datasets} & \textbf{Weight (\%)} \\
\midrule
Web & \href{https://huggingface.co/datasets/allenai/olmo-mix-1124}{DCLM}, \href{https://huggingface.co/datasets/HuggingFaceFW/fineweb-edu}{FineWeb-Edu}, \href{https://huggingface.co/datasets/allenai/dolma3_mix-6T-1025-7B}{Common Crawl} & 62.0\% \\
\midrule
Math & \href{https://huggingface.co/datasets/HuggingFaceTB/finemath/tree/main/finemath-3plus}{FineMath}, \href{https://huggingface.co/datasets/allenai/olmo-mix-1124}{OpenWebMath}, \href{https://huggingface.co/datasets/allenai/olmo-mix-1124}{Algebraic Stack}, \href{https://huggingface.co/datasets/nvidia/Llama-Nemotron-Post-Training-Dataset/blob/main/SFT/math/math_v1.1.jsonl}{Nemotron Math} & 11.6\% \\
\midrule
Code & \href{https://huggingface.co/datasets/allenai/olmo-mix-1124}{StarCoder}, \href{https://huggingface.co/datasets/allenai/dolma3_mix-6T-1025-7B}{Stack-Edu}, \href{https://huggingface.co/datasets/nvidia/Llama-Nemotron-Post-Training-Dataset/blob/main/SFT/code/code_v1.1.jsonl}{Nemotron Code} & 10.0\% \\
\midrule
Knowledge & \href{https://huggingface.co/datasets/allenai/dolmino-mix-1124}{FLAN}, \href{https://huggingface.co/datasets/HuggingFaceTB/smollm-corpus}{Cosmopedia}, \href{https://huggingface.co/datasets/allenai/olmo-mix-1124}{Wikipedia} & 10.0\% \\
\midrule
Science & \href{https://huggingface.co/datasets/allenai/dolma3_mix-6T-1025-7B}{OLMoCR Science PDFs}, \href{https://huggingface.co/datasets/allenai/olmo-mix-1124}{arXiv}, \href{https://huggingface.co/datasets/allenai/olmo-mix-1124}{peS2o}, \href{https://huggingface.co/datasets/nvidia/Llama-Nemotron-Post-Training-Dataset/blob/main/SFT/science/science.jsonl}{Nemotron Science}, \href{https://huggingface.co/datasets/facebook/natural_reasoning}{Natural Reasoning} & 6.4\% \\
\bottomrule
\end{tabular}
\end{table}

\paragraph{Mid-training data.} For mid-training to produce MobileMoE-Base, we shift the data distribution toward higher-quality, domain-specific sources while extending context length to 8,192. The mid-training mix combines two sources: (1) the \href{https://huggingface.co/datasets/allenai/dolma3_dolmino_mix-100B-1125}{Dolma3 Dolmino Mix}~\cite{olmo3}, a curated collection spanning synthetic math, code, QA, reasoning traces, instruction-following, high-quality web, and scientific PDFs; and (2) selected subsets from our pre-training data with code, math, and long-document sources upweighted for 8K context learning. While pre-training is web-heavy (62\%) for building general language modeling capacity, mid-training shifts toward higher-quality, domain-specific data: web is reduced (62\%$\to$9\%) while knowledge (10\%$\to$32\%), code (10\%$\to$22\%), and math (12\%$\to$21\%) are upweighted to strengthen downstream capabilities. This domain-concentrated mid-training further sharpens expert specialization in MoE, allowing routed experts to develop deeper expertise on domain-specific tokens. Table~\ref{tab:mt_data} summarizes the mid-training domain breakdown.

\begin{table}[!h]
\centering
\caption{\textbf{Mid-training data sources by domain.} Dolmino$^\dagger$ sources are from the \href{https://huggingface.co/datasets/allenai/dolma3_dolmino_mix-100B-1125}{Dolma3 Dolmino Mix}.}
\label{tab:mt_data}
\footnotesize
\setlength{\tabcolsep}{4pt}
\begin{tabular}{l p{0.72\linewidth} r}
\toprule
\textbf{Domain} & \textbf{Key Datasets} & \textbf{Weight (\%)} \\
\midrule
Knowledge & \href{https://huggingface.co/datasets/allenai/dolmino-mix-1124}{FLAN}, \href{https://huggingface.co/datasets/HuggingFaceTB/smollm-corpus}{Cosmopedia}, \href{https://huggingface.co/datasets/allenai/olmo-mix-1124}{Wiki}, \href{https://huggingface.co/datasets/allenai/dolma3_dolmino_mix-100B-1125}{Dolmino QA}$^\dagger$ & 32.2\% \\
\midrule
Code & \href{https://huggingface.co/datasets/allenai/olmo-mix-1124}{StarCoder}, \href{https://huggingface.co/datasets/allenai/dolma3_mix-6T-1025-7B}{Stack-Edu}, \href{https://huggingface.co/datasets/nvidia/Llama-Nemotron-Post-Training-Dataset/blob/main/SFT/code/code_v1.1.jsonl}{Nemotron Code}, \href{https://huggingface.co/datasets/allenai/dolma3_dolmino_mix-100B-1125}{Dolmino Code}$^\dagger$ & 22.1\% \\
\midrule
Math & \href{https://huggingface.co/datasets/HuggingFaceTB/finemath/tree/main/finemath-3plus}{FineMath}, \href{https://huggingface.co/datasets/allenai/olmo-mix-1124}{OpenWebMath}, \href{https://huggingface.co/datasets/allenai/olmo-mix-1124}{Algebraic Stack}, \href{https://huggingface.co/datasets/nvidia/Llama-Nemotron-Post-Training-Dataset/blob/main/SFT/math/math_v1.1.jsonl}{Nemotron Math}, \href{https://huggingface.co/datasets/allenai/dolma3_dolmino_mix-100B-1125}{Dolmino Math}$^\dagger$ & 21.2\% \\
\midrule
Science & \href{https://huggingface.co/datasets/allenai/dolma3_mix-6T-1025-7B}{OLMoCR Science PDFs}, \href{https://huggingface.co/datasets/allenai/olmo-mix-1124}{arXiv}, \href{https://huggingface.co/datasets/allenai/olmo-mix-1124}{peS2o}, \href{https://huggingface.co/datasets/nvidia/Llama-Nemotron-Post-Training-Dataset/blob/main/SFT/science/science.jsonl}{Nemotron Science} & 10.9\% \\
\midrule
Web & \href{https://huggingface.co/datasets/allenai/olmo-mix-1124}{DCLM}, \href{https://huggingface.co/datasets/HuggingFaceFW/fineweb-edu}{FineWeb-Edu}, \href{https://huggingface.co/datasets/allenai/dolma3_dolmino_mix-100B-1125}{Dolmino Web}$^\dagger$ & 8.5\% \\
\midrule
Instruction$^\dagger$ & \href{https://huggingface.co/datasets/allenai/dolma3_dolmino_mix-100B-1125}{Dolmino Instruction}$^\dagger$ & 4.4\% \\
\midrule
Reasoning$^\dagger$ & \href{https://huggingface.co/datasets/allenai/dolma3_dolmino_mix-100B-1125}{Dolmino Reasoning Traces}$^\dagger$ & 0.7\% \\
\bottomrule
\end{tabular}
\end{table}

\paragraph{Supervised fine-tuning (SFT) data.} We fine-tune MobileMoE-Base on a diverse mixture of open-licensed instruction-tuning datasets ($>80$M samples) from 28 public collections at 8K context length with sequence packing. The SFT mix spans multiple domains: math (30.4\%), general instruction/chat (25.4\%), code (22.1\%), safety (9.4\%), science/knowledge (7.7\%), tool use (3.9\%), and reasoning (1.1\%), covering a broad range of capabilities for instruction-following. Each dataset is assigned a sampling weight $w_i = \max(1, \lfloor n_i / N \times 100 \rceil)$, where $n_i$ is the dataset sample count and $N$ is the total sample count, which assigns sampling probability proportional to dataset size while guaranteeing that small but important domains (e.g., safety, tool use) receive sufficient representation via the $\max(1, \cdot)$ floor. Table~\ref{tab:sft_data} lists the SFT data sources by domain.

\begin{table}[!h]
\centering
\caption{\textbf{SFT data sources by domain} ($>80$M samples). Dataset names are linked to their repositories.}
\label{tab:sft_data}
\footnotesize
\setlength{\tabcolsep}{3pt}
\begin{tabular}{l p{0.62\linewidth} r r}
\toprule
\textbf{Domain} & \textbf{Key Datasets} & \textbf{\# Samples} & \textbf{Weight (\%)} \\
\midrule
Math & \href{https://huggingface.co/datasets/nvidia/Llama-Nemotron-Post-Training-Dataset}{Nemotron PTD}, \href{https://huggingface.co/datasets/nvidia/Nemotron-Post-Training-Dataset-v1}{Nemotron SFT}, \href{https://huggingface.co/datasets/nvidia/OpenMathInstruct-2}{OpenMathInstruct}, \href{https://huggingface.co/datasets/nvidia/Puzzle-KD-Nemotron-Post-Training-Dataset-v2}{Puzzle-KD}, \href{https://huggingface.co/datasets/allenai/Dolci-Instruct-SFT}{Dolci}, \href{https://huggingface.co/datasets/HuggingFaceTB/smoltalk}{SmolTalk} & 39.2M & 30.4\% \\
\midrule
Code & \href{https://huggingface.co/datasets/nvidia/OpenCodeGeneticInstruct}{OpenCodeGeneticInstruct}, \href{https://huggingface.co/datasets/nvidia/OpenCodeInstruct}{OpenCodeInstruct}, \href{https://huggingface.co/datasets/nvidia/Nemotron-Post-Training-Dataset-v1}{Nemotron SFT}, \href{https://huggingface.co/datasets/glaiveai/glaive-code-assistant}{Glaive}, \href{https://huggingface.co/datasets/allenai/Dolci-Instruct-SFT}{Dolci}, \href{https://huggingface.co/datasets/nvidia/Nemotron-SWE-v1}{Nemotron-SWE} & 23.1M & 22.1\% \\
\midrule
General Chat & \href{https://huggingface.co/datasets/HuggingFaceTB/smoltalk}{SmolTalk}, \href{https://huggingface.co/datasets/nvidia/Nemotron-Post-Training-Dataset-v1}{Nemotron SFT}, \href{https://huggingface.co/datasets/allenai/Dolci-Instruct-SFT}{Dolci}, \href{https://huggingface.co/datasets/allenai/tulu-3-sft-mixture}{Tulu-3 SFT}, \href{https://huggingface.co/datasets/openbmb/UltraChat}{UltraChat}, \href{https://huggingface.co/datasets/mlabonne/open-perfectblend}{open-perfectblend}, \href{https://huggingface.co/datasets/nvidia/Daring-Anteater}{Daring-Anteater}, \href{https://huggingface.co/datasets/allenai/dolmino-mix-1124}{FLAN}, \href{https://huggingface.co/datasets/microsoft/orca-agentinstruct-1M-v1}{Orca-AgentInstruct}, \href{https://huggingface.co/datasets/nvidia/Retrieval-Synthetic-NVDocs-v1}{Retrieval-NVDocs}, \href{https://huggingface.co/datasets/nvidia/HelpSteer2}{HelpSteer}, \href{https://huggingface.co/datasets/OpenAssistant/oasst2}{OASST2} & 9.8M & 25.4\% \\
\midrule
Science/Knowledge & \href{https://huggingface.co/datasets/nvidia/OpenScience}{OpenScience}, \href{https://huggingface.co/datasets/nvidia/Llama-Nemotron-Post-Training-Dataset}{Nemotron PTD}, \href{https://huggingface.co/datasets/nvidia/Nemotron-Post-Training-Dataset-v2}{Nemotron SFT v2}, \href{https://huggingface.co/datasets/allenai/Dolci-Instruct-SFT}{Dolci} & 7.6M & 7.7\% \\
\midrule
Reasoning & \href{https://huggingface.co/datasets/Alibaba-PAI/OmniThought-0528}{OmniThought}, \href{https://huggingface.co/datasets/HuggingFaceTB/smoltalk}{SmolTalk} & 0.8M & 1.1\% \\
\midrule
Tool Use & \href{https://huggingface.co/datasets/nvidia/Nemotron-Post-Training-Dataset-v1}{Nemotron SFT}, \href{https://huggingface.co/datasets/Team-ACE/ToolACE}{ToolACE}, \href{https://huggingface.co/datasets/argilla/Synth-APIGen-v0.1}{Synth-APIGen}, \href{https://huggingface.co/datasets/HuggingFaceTB/smoltalk}{SmolTalk} & 0.5M & 3.9\% \\
\midrule
Safety & \href{https://huggingface.co/datasets/nvidia/Llama-Nemotron-Post-Training-Dataset}{Nemotron PTD}, \href{https://huggingface.co/datasets/nvidia/Nemotron-Safety-Guard-Dataset-v3}{Safety-Guard}, \href{https://huggingface.co/datasets/nvidia/Nemotron-PII}{PII}, \href{https://huggingface.co/datasets/nvidia/Aegis-AI-Content-Safety-Dataset-2.0}{Aegis}, \href{https://huggingface.co/datasets/allenai/Dolci-Instruct-SFT}{Dolci} & 0.5M & 9.4\% \\
\bottomrule
\end{tabular}
\end{table}

\section{Evaluation Details}
\label{app:eval}

\subsection{Evaluation setup of base and instruct models on foundational competencies}
\label{app:eval-foundational}

\hypertarget{app:eval-base}{}%
We evaluate both pre-trained (base) and instruction-tuned (SFT) models on 14 foundational benchmarks using the Language Model Evaluation Harness (\href{https://github.com/EleutherAI/lm-evaluation-harness}{lm-eval})~\cite{eval-harness} with the \href{https://github.com/vllm-project/vllm}{vLLM} backend (\texttt{dtype=auto}, resolved to bfloat16 at model precision; \texttt{add\_bos\_token=True}). All runs use \texttt{--batch\_size auto:16} and greedy decoding (\texttt{temperature=0}, \texttt{do\_sample=false}) for deterministic results. Instruct models are evaluated in non-thinking mode. We follow standard few-shot settings for both base and instruction models; per-task configurations are summarized in Table~\ref{tab:eval-config}.

\begin{table}[!h]
\centering
\caption{\textbf{Evaluation benchmark configurations on foundational competencies (base and instruct models).} Evaluation is using \href{https://github.com/EleutherAI/lm-evaluation-harness}{lm-eval}.}
\label{tab:eval-config}
\footnotesize
\setlength{\tabcolsep}{3pt}
\begin{tabular}{lllll}
\toprule
\textbf{Category} & \textbf{Benchmark} & \textbf{Task name} & \textbf{n-shot} & \textbf{Metric} \\
\midrule
\multirow{4}{*}{Commonsense Reasoning} & HellaSwag & \texttt{hellaswag} & 0 & \texttt{acc\_norm} \\
 & PIQA & \texttt{piqa} & 0 & \texttt{acc\_norm} \\
 & SIQA & \texttt{social\_iqa} & 0 & \texttt{acc} \\
 & WinoGrande & \texttt{winogrande} & 0 & \texttt{acc} \\
\midrule
\multirow{3}{*}{Knowledge} & MMLU & \texttt{mmlu} & 5 & \texttt{acc} \\
 & NQ & \texttt{nq\_open} & 5 & \texttt{exact\_match} \\
 & TQA & \texttt{triviaqa} & 5 & \texttt{exact\_match} \\
\midrule
\multirow{3}{*}{Science} & ARC-C & \texttt{arc\_challenge} & 25 & \texttt{acc\_norm} \\
 & ARC-E & \texttt{arc\_easy} & 0 & \texttt{acc\_norm} \\
 & OBQA & \texttt{openbookqa} & 0 & \texttt{acc\_norm} \\
\midrule
\multirow{2}{*}{Reading} & BoolQ & \texttt{boolq} & 0 & \texttt{acc} \\
 & DROP & \texttt{drop} & 3 & \texttt{f1} \\
\midrule
\multirow{2}{*}{Reasoning} & BBH & \texttt{leaderboard\_bbh} & 3 & \texttt{acc\_norm} \\
 & GSM8K & \texttt{gsm8k\_cot} & 8 & \texttt{exact\_match,flexible-extract} \\
\bottomrule
\end{tabular}
\end{table}

\subsection{Evaluation setup of instruct models on advanced competencies}
\label{app:eval-advanced}

\hypertarget{app:eval-instruct}{}%
For instruct-tuned models, we evaluate 8 advanced benchmarks spanning four capability axes: math, code, instruction following, and harder knowledge \& reasoning. We use \href{https://github.com/EleutherAI/lm-evaluation-harness}{lm-eval}~\cite{eval-harness} for MATH500, GSM-Plus, HumanEval MBPP, IFEval, and GPQA Diamond, and the official packages \href{https://github.com/TIGER-AI-Lab/MMLU-Pro}{TIGER-AI-Lab/MMLU-Pro}~\cite{mmlupro} and \href{https://github.com/allenai/IFBench}{allenai/IFBench}~\cite{ifbench} for MMLU-Pro and IFBench. For lm-eval benchmarks, we use the \href{https://github.com/vllm-project/vllm}{vLLM} backend (\texttt{dtype=auto}, resolved to bfloat16 at model precision; \texttt{add\_bos\_token=True}), \texttt{--batch\_size auto:16}, and greedy decoding (\texttt{temperature=0}, \texttt{do\_sample=false}) for deterministic results. Instruct models are evaluated in non-thinking mode. For MMLU-Pro and IFBench, we use the default settings from official packages. Per-task settings are summarized in Table~\ref{tab:eval-config-instruct}.

\begin{table}[!h]
\centering
\caption{\textbf{Evaluation benchmark configurations on advanced competencies (instruct models).} Generation tasks use chat templates and greedy decoding ($T=0$) unless noted; loglikelihood tasks do not generate.}
\label{tab:eval-config-instruct}
\footnotesize
\setlength{\tabcolsep}{3pt}
\begin{tabular}{lllccl}
\toprule
\textbf{Category} & \textbf{Benchmark} & \textbf{Task name / Script (Package)} & \textbf{Shots} & \textbf{Chat} & \textbf{Metric} \\
\midrule
\multirow{2}{*}{Math} & MATH500 & \texttt{minerva\_math500} (lm-eval) & 4 & No & \texttt{math\_verify} \\
 & GSM-Plus & \texttt{gsm\_plus} (lm-eval) & 5 & Yes & \texttt{flexible-extract} \\
\midrule
\multirow{2}{*}{Code} & HumanEval$_{p@1}$ & \texttt{humaneval\_instruct} (lm-eval) & 0 & Yes & pass@1 (greedy) \\
 & MBPP & \texttt{mbpp\_instruct} (lm-eval) & 3 & Yes & pass@1 (greedy) \\
\midrule
\multirow{2}{*}{IF} & IFEval & \texttt{ifeval} (lm-eval) & 0 & Yes & Avg(strict/loose $\times$ prompt/inst) \\
 & IFBench & \texttt{ifbench\_generate.py} (AllenAI) & 0 & Yes & Avg(strict/loose $\times$ prompt/inst) \\
\midrule
Knowledge & MMLU-Pro & \texttt{evaluate\_from\_local.py} (TIGER-Lab) & 5 (CoT) & No & acc (regex \texttt{answer is (X)}) \\
Reasoning & GPQA Diamond & \texttt{gpqa\_diamond\_zeroshot} (lm-eval) & 0 & No & acc \\
\bottomrule
\end{tabular}
\end{table}

\subsection{Baseline Model Sources}
\label{app:baselines}

All baseline models are publicly available. We list the HuggingFace model identifiers used in our evaluation for both base (pre-trained) and instruct (SFT) models in Table \ref{tab:baselines}.

\begin{table}[!h]
\centering
\caption{\textbf{Baseline model sources.} \href{https://huggingface.co}{HuggingFace} identifiers are given for all base and instruct models.}
\label{tab:baselines}
\footnotesize
\setlength{\tabcolsep}{3pt}
\begin{tabular}{llll}
\toprule
\textbf{Model} & \textbf{$N_{\text{act}}$} & \textbf{Base (PT)} & \textbf{Instruct (SFT)} \\
\midrule
Gemma~3 270M & 270M & \texttt{google/gemma-3-270m} & \texttt{google/gemma-3-270m-it} \\
SmolLM2 360M & 362M & \texttt{HuggingFaceTB/SmolLM2-360M} & \texttt{HuggingFaceTB/SmolLM2-360M-Instruct} \\
Granite-3.0-1B & 400M/1.3B & -- & \texttt{ibm-granite/granite-3.0-1b-a400m-instruct} \\
Qwen3.5 0.8B & 749M & \texttt{Qwen/Qwen3.5-0.8B-Base} & \texttt{Qwen/Qwen3.5-0.8B} \\
Granite-3.0-3B & 800M/3.3B & -- & \texttt{ibm-granite/granite-3.0-3b-a800m-instruct} \\
Gemma~3 1B & 1.0B & \texttt{google/gemma-3-1b-pt} & \texttt{google/gemma-3-1b-it} \\
MobileLLM-Pro & 1.1B & \texttt{facebook/MobileLLM-Pro-base} & \texttt{facebook/MobileLLM-Pro} \\
Llama~3.2 1B & 1.2B & \texttt{meta-llama/Llama-3.2-1B} & \texttt{meta-llama/Llama-3.2-1B-Instruct} \\
OLMoE-1B-7B & 1.3B/6.9B & \texttt{allenai/OLMoE-1B-7B-0924} & \texttt{allenai/OLMoE-1B-7B-0924-Instruct} \\
OLMo~2 1B & 1.5B & \texttt{allenai/OLMo-2-0425-1B} & \texttt{allenai/OLMo-2-0425-1B-Instruct} \\
SmolLM2 1.7B & 1.7B & \texttt{HuggingFaceTB/SmolLM2-1.7B} & \texttt{HuggingFaceTB/SmolLM2-1.7B-Instruct} \\
Qwen3.5 2B & 1.9B & \texttt{Qwen/Qwen3.5-2B-Base} & \texttt{Qwen/Qwen3.5-2B} \\
Jet-Nemotron-2B & 2.0B & -- & \texttt{jet-ai/Jet-Nemotron-2B} \\
Jet-Nemotron-4B & 4.0B & -- & \texttt{jet-ai/Jet-Nemotron-4B} \\
\bottomrule
\end{tabular}
\end{table}

\subsection{MMLU-Pro and GPQA Diamond per-protocol ablation}
\label{app:eval-variants}

\textbf{Evaluation protocols.} We compare MMLU-Pro and GPQA Diamond evaluation protocols in Table~\ref{tab:eval-variants}. We report three MMLU-Pro (5-shot) variants: (1) the official package \href{https://github.com/TIGER-AI-Lab/MMLU-Pro}{TIGER-AI-Lab/MMLU-Pro}, which uses CoT generation, no chat template, and a subject-specific prompt; (2) \href{https://github.com/EleutherAI/lm-evaluation-harness}{lm-eval} (LLH), which evaluates with deterministic loglikelihood scoring; and (3) \href{https://github.com/EleutherAI/lm-evaluation-harness}{lm-eval} (Chat), which evaluates with CoT generation and chat template. For GPQA Diamond (0-shot), we report two variants: (1) \href{https://github.com/EleutherAI/lm-evaluation-harness}{lm-eval} (LLH), multiple-choice loglikelihood scoring, and (2) \href{https://github.com/EleutherAI/lm-evaluation-harness}{lm-eval} (Chat), CoT generation with chat template.

\begin{table}[!t]
\centering
\caption{\textbf{MMLU-Pro and GPQA Diamond per-protocol ablation.} 5-shot MMLU-Pro and 0-shot GPQA Diamond across 13 instruct-tuned baselines under multiple evaluation protocols from \href{https://github.com/TIGER-AI-Lab/MMLU-Pro}{TIGER-AI-Lab/MMLU-Pro} and \href{https://github.com/EleutherAI/lm-evaluation-harness}{lm-eval}. Main-body Table~\ref{tab:sft-appendix} reports MMLU-Pro with \href{https://github.com/TIGER-AI-Lab/MMLU-Pro}{TIGER-AI-Lab/MMLU-Pro} and GPQA Diamond with \href{https://github.com/EleutherAI/lm-evaluation-harness}{lm-eval} (LLH). LLH: loglikelihood.}
\label{tab:eval-variants}
\footnotesize
\setlength{\tabcolsep}{4pt}
\resizebox{0.8\linewidth}{!}{%
\begin{tabular}{@{}lc ccc cc@{}}
\toprule
& & \multicolumn{3}{c}{\textbf{MMLU-Pro} (5-shot)} & \multicolumn{2}{c}{\textbf{GPQA Diamond} (0-shot)} \\
\cmidrule(lr){3-5} \cmidrule(lr){6-7}
\textbf{Model} & $N_{\text{act}}/N_{\text{total}}$ & \href{https://github.com/TIGER-AI-Lab/MMLU-Pro}{TIGER-AI-Lab} & \href{https://github.com/EleutherAI/lm-evaluation-harness}{lm-eval} (LLH) & \href{https://github.com/EleutherAI/lm-evaluation-harness}{lm-eval} (Chat) & \href{https://github.com/EleutherAI/lm-evaluation-harness}{lm-eval} (LLH) & \href{https://github.com/EleutherAI/lm-evaluation-harness}{lm-eval} (Chat) \\
\midrule
Gemma~3 270M & 270M & 11.3 & 11.3 & \phantom{0}0.0 & 25.8 & 19.2 \\
SmolLM2 360M & 362M & 12.0 & 10.9 & \phantom{0}9.9 & 25.8 & \textbf{26.8} \\
\textbf{MobileMoE-S} & \textbf{272M/1.3B} & \textbf{18.2} & \textbf{18.1} & \textbf{14.0} & \textbf{27.8} & 21.2 \\
\midrule
Qwen3.5 0.8B & 749M & 24.0 & 24.3 & \textbf{31.2} & \textbf{26.3} & 20.2 \\
\textbf{MobileMoE-M} & \textbf{528M/2.8B} & \textbf{28.3} & \textbf{25.0} & 26.1 & 24.8 & \textbf{22.7} \\
\midrule
Gemma~3 1B & 1.0B & 16.1 & 15.1 & \phantom{0}0.0 & 25.3 & 25.3 \\
MobileLLM-Pro & 1.1B & 10.9 & 11.5 & 15.5 & 23.2 & 18.2 \\
Llama~3.2 1B & 1.2B & 20.8 & 19.1 & 13.0 & 28.8 & 20.7 \\
OLMo~2 1B & 1.5B & 16.0 & 15.9 & 15.0 & 29.8 & 26.3 \\
SmolLM2 1.7B & 1.7B & 19.8 & 20.6 & 20.9 & 29.8 & 20.7 \\
Qwen3.5 2B & 1.9B & \textbf{38.8} & \textbf{31.5} & \textbf{48.9} & \textbf{34.3} & \textbf{43.9} \\
OLMoE-1B-7B & 1.3B/6.9B & 19.5 & 18.4 & 18.7 & 24.2 & 24.2 \\
\textbf{MobileMoE-L} & \textbf{922M/5.3B} & 34.0 & 29.3 & 33.9 & 33.8 & 26.8 \\
\bottomrule
\end{tabular}
}%
\end{table}

\textbf{Analysis on evaluation protocols.} As Table~\ref{tab:eval-variants} shows, applying the chat template hurts most baselines: MMLU-Pro with \href{https://github.com/EleutherAI/lm-evaluation-harness}{lm-eval} (Chat) regresses on 9 of 13 models vs. the results with original evaluation package \href{https://github.com/TIGER-AI-Lab/MMLU-Pro}{TIGER-AI-Lab} (e.g., Gemma~3 1B $0.0$ vs.\ $16.1$, Llama~3.2 1B $13.0$ vs.\ $20.8$) while substantially benefiting only Qwen3.5 2B ($48.9$ vs.\ $38.8$). Similarly, GPQA Diamond with \href{https://github.com/EleutherAI/lm-evaluation-harness}{lm-eval} (Chat) shows the same pattern, helping Qwen3.5 2B ($+9.6$) while degrading 9 of 13 models. To avoid the bias and collapse brought by model-specific chat templates, Table~\ref{tab:sft-appendix} uses the no-chat protocols: \href{https://github.com/TIGER-AI-Lab/MMLU-Pro}{TIGER-AI-Lab} for MMLU-Pro, \href{https://github.com/EleutherAI/lm-evaluation-harness}{lm-eval} (LLH) for GPQA Diamond, which give the better score for most models and ensure a consistent comparison among all models.

\section{Quantitative Analysis}
\label{app:quantitative}

\begin{figure}[!h]
\centering
\includegraphics[width=0.75\linewidth]{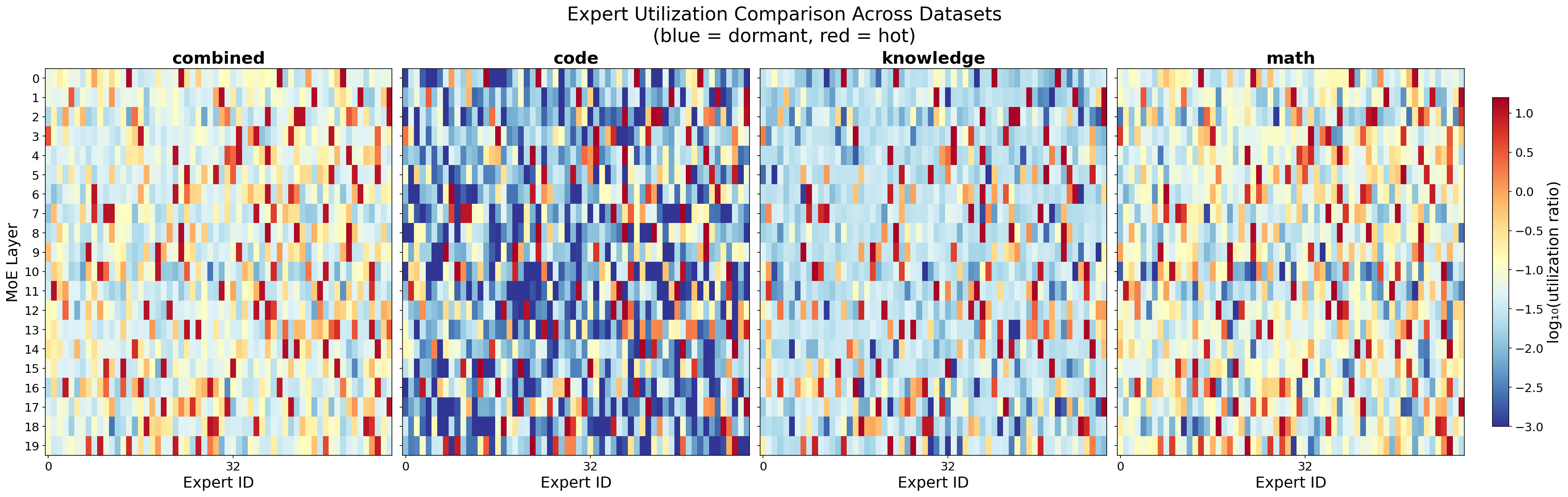}\\[2pt]
{\small (a) Pre-training (PT) stage}\\[6pt]
\includegraphics[width=0.75\linewidth]{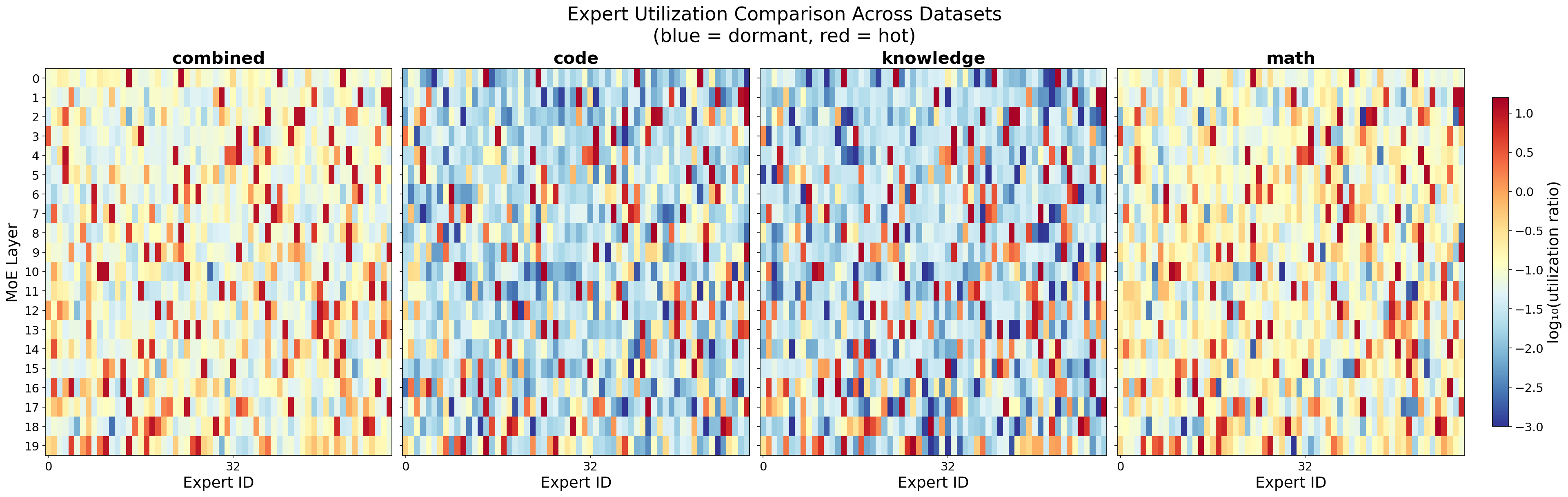}\\[2pt]
{\small (b) Mid-training (MT) stage}\\[6pt]
\includegraphics[width=0.75\linewidth]{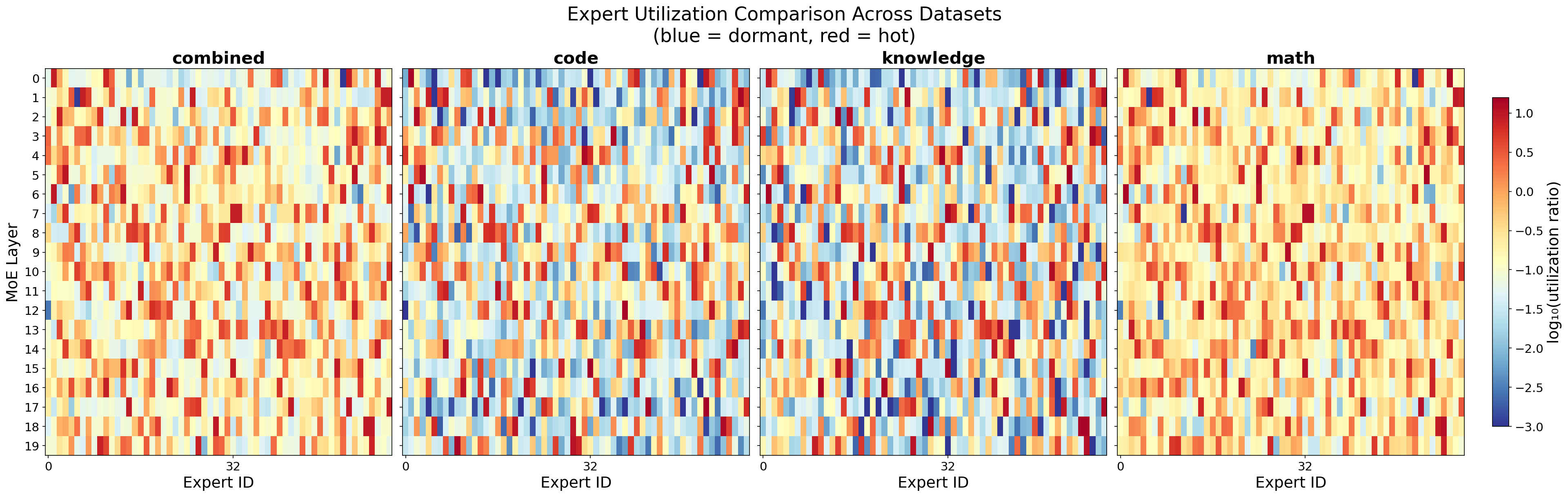}\\[2pt]
{\small (c) Supervised fine-tuning (SFT) stage}
\caption{\textbf{MobileMoE expert utilization heatmaps across training stages and domains.} Each heatmap shows per-layer (rows) per-expert (columns) activation utilization on the log$_{10}$ scale (blue: dormant, red: hot) for MobileMoE-S evaluated on three downstream domains. Two patterns emerge: (i) different domains activate distinct expert subsets (cross-task specialization), and (ii) expert utilization broadens progressively from PT through MT to SFT.}
\label{fig:expert-heatmap}
\end{figure}

\textbf{Visualization of MobileMoE expert utilization.} We visualize per-layer expert utilization patterns across downstream tasks: code, math, knowledge for MobileMoE-S after pre-training (PT), mid-training (MT), and supervised fine-tuning (SFT) stages, where lower utilization (blue) indicates dormant experts and higher utilization (red) indicates frequently activated experts. Figure~\ref{fig:expert-heatmap} reveals these patterns: \textbf{(1)} \textit{Expert specialization differs across tasks:} on different domains, different subsets of experts are activated, suggesting the 60 fine-grained experts specialize across distinct domains. \textbf{(2)} \textit{Expert utilization broadens through training:} at PT, fewer experts are highly utilized; through MT and SFT, more experts are progressively activated, indicating that downstream training broadens expert utilization while maintaining cross-task specialization.

\begin{figure}[!h]
\centering
\includegraphics[width=0.85\linewidth]{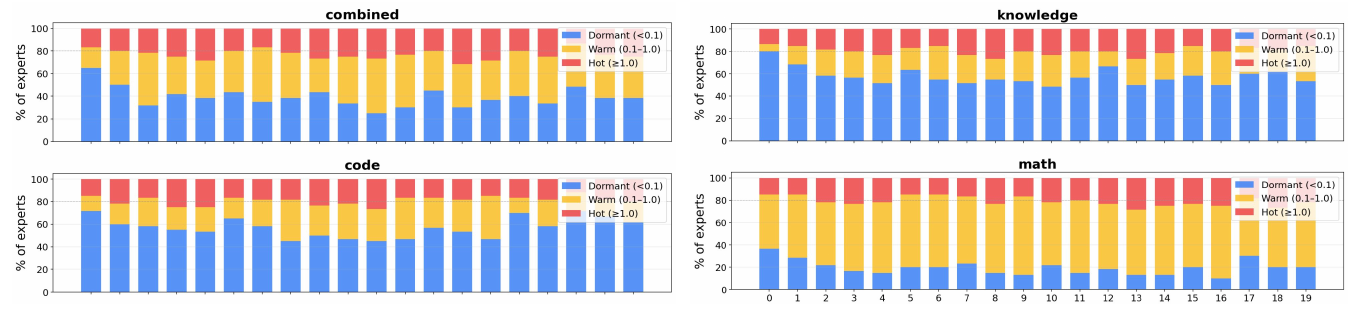}
\caption{\textbf{Distribution of expert utilization across tasks: math, code, knowledge, combined (all tasks).} Statistics of expert utilization ratios (log$_{10}$ scale) are shown across all MoE layers for MobileMoE-S (post SFT). }
\label{fig:expert-distribution}
\end{figure}

\textbf{Expert utilization statistics.} The distribution of expert utilization ratios varies across downstream tasks (Figure~\ref{fig:expert-distribution}): math activates a broader set of experts, while code or knowledge tasks concentrate on a narrower subset. This task-dependent sparsity indicates that not every expert weight needs to be loaded at inference, opening a path to save on-device memory via selective expert loading or task-conditional pruning.

\end{document}